\documentclass{article}

\usepackage{microtype}
\usepackage{graphicx}
\usepackage{subcaption}
\usepackage{booktabs} % for professional tables
\usepackage{verbatim}

\usepackage{hyperref}

\usepackage[accepted]{icml2024}
\usepackage{amsmath}
\usepackage{amssymb}
\usepackage{mathtools}
\usepackage{amsthm}

\usepackage[capitalize,noabbrev]{cleveref}

\theoremstyle{plain}

\theoremstyle{definition}

\theoremstyle{remark}

\usepackage[textsize=tiny]{todonotes}

\usepackage{xcolor}

\definecolor{pastel-red}{HTML}{FF9AA2}
\definecolor{pastel-green}{HTML}{B5EAD7}
\definecolor{pastel-blue}{HTML}{A7DBD8}
\definecolor{pastel-teal}{HTML}{9AD9DB}
\definecolor{pastel-yellow}{HTML}{FFFFCC}
\definecolor{pastel-purple}{HTML}{D3C0F9}
\definecolor{pastel-orange}{HTML}{FFC2A2}
\definecolor{pastel-pink}{HTML}{FFB7B2}
\definecolor{pastel-cyan}{HTML}{A1E8E2}
\definecolor{pastel-lavender}{HTML}{E0BBE4}

\usepackage{amsfonts}
\usepackage{color}

\usepackage{xspace}

\icmltitlerunning{Reasoning in Token Economies}

\begin{document}

\twocolumn[
\icmltitle{Reasoning in Token Economies: Budget-Aware Evaluation of \\
                                    LLM Reasoning Strategies}



\icmlsetsymbol{intern}{*}
\icmlsetsymbol{nv}{\S}

\begin{icmlauthorlist}
\icmlauthor{Junlin Wang}{duke,intern}
\icmlauthor{Siddhartha Jain}{nvidia,nv}
\icmlauthor{Dejiao Zhang}{aws}
\icmlauthor{Baishakhi Ray}{aws}
\icmlauthor{Varun Kumar}{aws}
\icmlauthor{Ben Athiwaratkun}{together}
\end{icmlauthorlist}

\icmlaffiliation{aws}{AWS AI Labs}
\icmlaffiliation{duke}{Duke University}
\icmlaffiliation{together}{Together AI}
\icmlaffiliation{nvidia}{Nvidia}
\icmlcorrespondingauthor{Junlin Wang}{junlin.wang2@duke.edu}
\icmlcorrespondingauthor{Siddhartha Jain}{tmfs10@gmail.com}

\icmlkeywords{Machine Learning, ICML}

\vskip 0.3in
]


\printAffiliationsAndNotice{\textsuperscript{*}Work conducted during an internship at Amazon. \S Work done while at Amazon.} 
% \printAffiliationsAndNotice{\textsuperscript{*}Work conducted during an internship at Amazon} 
% \printAffiliationsAndNotice{\textsuperscript{^}Work done while at Amazon}
% leave blank if no need to mention equal contribution

\begin{abstract}
A diverse array of reasoning strategies has been proposed to elicit the capabilities of large language models.
However, in this paper, we point out that traditional evaluations which focus solely on performance metrics miss a key factor: the increased effectiveness due to additional compute.
By overlooking this aspect, a skewed view of strategy efficiency is often presented.
This paper introduces a framework that incorporates the compute budget into the evaluation, providing a more informative comparison that takes into account both performance metrics and computational cost.
%While it is expected that the reasoning performance improves with respect to compute budget, we point out an important observation that more complex reasoning methods perform well due to the associated compute rather than the ingenuity of such methods.
In this budget-aware perspective, we find that complex reasoning strategies often don't surpass simpler baselines purely due to algorithmic ingenuity, but rather due to the larger computational resources allocated. When we provide a simple baseline like chain-of-thought self-consistency with comparable compute resources, it frequently outperforms reasoning strategies proposed in the literature.
In this scale-aware perspective, we find that unlike self-consistency, certain strategies such as multi-agent debate or Reflexion can become worse if more compute budget is utilized. 
\end{abstract}

\begin{figure*}[ht!]
% \vspace{-0.3cm}
\begin{center}
\centering
\includegraphics[width=0.9\textwidth]{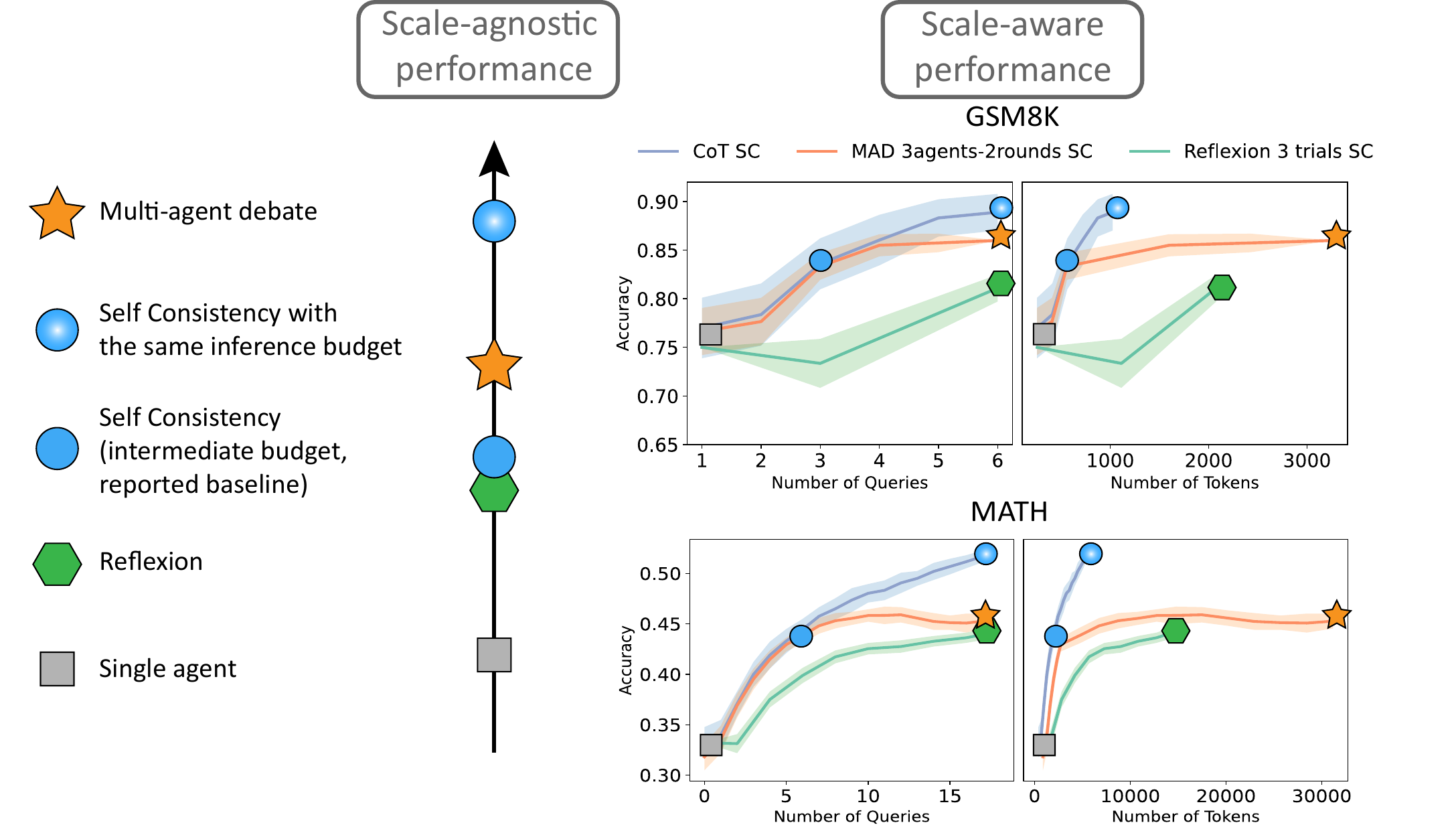}
% \begin{subfigure}[b]{0.47\textwidth}
% \includegraphics[width=1\textwidth]{figures/figures_gsm8k/6samples_100_figure1a_ai2.pdf}
% % \includegraphics[width=0.48 \textwidth]{figures/figures_gsm8k/total_tokens_6samples_100_figure2a.pdf}

% \caption{ 
% GSM8K
% } \label{fig:scale_aware}
% \end{subfigure}
% \begin{subfigure}[b]{0.47\textwidth}
% \includegraphics[width=1\textwidth]{figures/figures_MATH/figure1b_ai2.pdf} 
% % \includegraphics[width=0.45 \textwidth]{figures/figures_HotpotQA/queries_19samples_500_figure2b.pdf} 
% % \includegraphics[width=0.45 \textwidth]{figures/figures_HotpotQA/total_tokens_19samples_500_figure2b.pdf}
% \caption{
% MATH
% } \label{fig:oracle_nonoracle}
% \end{subfigure}
\vspace{-0.1cm}
\end{center}
\caption{
(1) Comparison of reasoning approaches multi-agent debate (MAD) against the SC baseline, considering both scale-agnostic and scale-aware evaluation, with published scores and our reproductions on the GSM8K and MATH dataset. The scale-aware evaluation furnishes more comprehensive insights into the influence of scale on reasoning strategies and offers a fairer method of comparison. 
(2) The scale-aware comparison between Reflexion and SC also illustrates the artifact of scale on performance. 
For both datasets, we show both budgets, the number of total tokens, and the number of queries. All results were obtained from GPT-3.5.
} \label{fig:inference_scale}
%\ben{how about top row versus bottom row -- using (1) and (2) may be confusing}
\end{figure*}

\section{Introduction}
\label{sec:intro}

The arena of large language models (LLMs) such as GPT-4 \cite{OpenAI2023GPT4,touvron2023llama,team2023gemini,jiang2023mistral} has seen a proliferation of diverse reasoning strategies. However, comparing these strategies fairly and comprehensively has proven to be a challenging task due to their varied computational requirements. For instance, strategies like tree of thoughts (ToT) necessitate branching out into multiple sequences and incorporating self-evaluation, making them more compute-intensive than others. Therefore, an evaluation framework that only accounts for performance metrics may miss crucial practical factors such as computational cost.

In this paper, we propose the inclusion of the compute budget into the performance measurement of different reasoning strategies. This budget-aware comparison yields a more balanced perspective on the effectiveness of reasoning strategies, accounting for both the quality of the output and the computational resources expended.

Our empirical research uncovers a significant correlation between the performance and the compute budget. We find that a straightforward baseline strategy, chain-of-thought reasoning coupled with self-consistency, can be remarkably competitive. When scaled to match the compute resources of more sophisticated methods such as Multi-Agent Debate (MAD) \citep{liang2023encouraging}, Reflexion \citep{shinn2023reflexion}, Plan and Solve \cite{wang2023planandsolve}, Least to Most Prompting \cite{zhou2022leasttomost}, Progressive Hint Prompting \cite{zheng2023progressivehint}, this baseline strategy often outperforms them in achieving the best trade-off between performance and budget. We further investigate the reasons behind the gap from simple CoT SC and other reasoning strategies by providing both empirical and theoretical evidence.

Then we scrutinize the influence of two specific types of budgets on performance: (1) the answer generation budget, and (2) the evaluation budget. {Our findings indicate that self-evaluation performance is really dependent on models and datasets. Moreover, we identify a strong correlation between the calibration via a correctness prediction proxy and the success of reasoning strategies that leverage self-evaluation.

%Based on this, we also propose Self-Confidence-weighted self-evaluation $SC^2$ as another simple yet effective baseline for math reasoning tasks. 

This work provides a robust framework for comparing a wide array of reasoning strategies and illuminates the significance of self-evaluation in these models. We hope this sets the stage for more focused research on efficient budget utilization and paves the way for the development of even more effective reasoning strategies.

Concretely, our contributions are 
\begin{itemize}
    \item We introduce a budget-aware evaluation framework spanning three dimensions: queries, tokens, and monetary cost, advocating for the token-based metric as the most holistic. This metric adeptly captures both the latency and financial implications of computational tasks. 
    \item We present a comprehensive evaluation of seven LLM reasoning strategies across five datasets using five models including GPT-4. Our analysis reveals that traditional evaluation metrics often overlook a critical aspect: the performance gains achievable through additional computational resources. This observation is strongly supported by CoT SC matches or even exceeds more complex strategies in effectiveness.
    \item We explore the dynamics of reasoning strategies, highlighting that MAD underperforms as diversity diminishes with each round. Conversely, Self-Consistency excels due to the independence of samples boosting diversity and its effectiveness in scenarios where the likelihood of being correct exceeds 50\%.
    \item We conduct ablation studies on ToT and Reflexion by segregating the budget into answer generation and evaluation budgets. We found that self-evaluation is promising at increasing performance while being cost-effective but currently LLM can't self-evaluate well.
\end{itemize}

% \begin{itemize}

% \item Calibration: another evidence that doing actual evaluation, answers that are majority tend to be more correct. \ben{can we show negative results for calibration?}
% \ben{look at the math behind self-consistency --
% The performance for certain datasets plateau because it's a mixture: for some problem, the score converges to 1. for some problem, the score converges to 0. (maybe make sure this is the case)
% Ben -- is this true? if we only pick one problem, is self-consistency over a given problem either 0 or 1? I think so? 
% }

% \item 
% \ben{ mention results of mixtral etc }
% \ben{mention that we have done implemented additional reasoning strategies clearly in main paper and link back.
% Mention something strong like: 'self-consistency outperforms most strategies when compared in an budget-aware manner, except only in certain cases such as 'xxx' and 'xxxx' for instance'
% }

% \end{itemize}

\section{Related Work}\label{sec:related}

\begin{figure*}[ht!]
\begin{center}
\vspace{-0.2cm}
%\framebox[4.0in]{$\;$}
%\fbox{\rule[-.5cm]{0cm}{4cm} \rule[-.5cm]{4cm}{0cm}}
\includegraphics[width=1\textwidth]{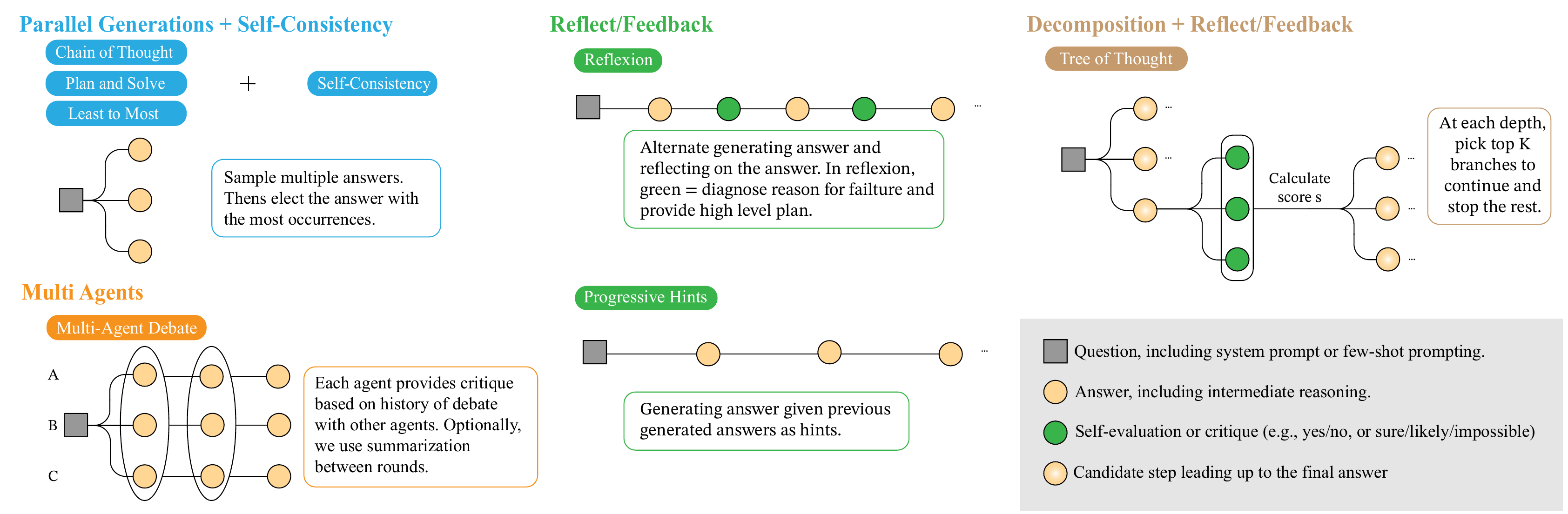}
\end{center}
\vspace*{-3mm}
\caption{Overview of reasoning strategies. Green cell indicates question prompt, including system prompt and few-shot prompting. The orange cell indicates the answer. Blue cell indicates evaluation or critique. } \label{fig:overview_reasoning}
\end{figure*}

\subsection{Reasoning strategies for LLMs}

There has been a flurry of activity to use language models for generating effective reasoning and planning strategies. An early work in the area was to prompt the language model to generate its Chain-of-Thought (CoT)~\citep{wei2022chain} when solving a problem which led to significant improvements in the model’s problem-solving abilities. Later work has involved prompting the language model to come up with its plan for solving the problem before trying to solve it~\citep{jiang2023self}, using chain-of-thought to solve a problem and then asking the model to critique and revise its solution (feedback)~\citep{madaan2023self, scheurer2023training, chen2023improving, bai2022constitutional,  kim2023language, shinn2023reflexion}, generating multiple chain-of-thoughts and combining them using LLM~\citep{yoran2023answering}, setting up a tree search for chain-of-thought instead of sampling a single linear chain-of-thought (\textit{Tree of Thoughts} - ToT)~\citep{yao2023tree}, aggregating LLM generated feedback for multiple prompts and their solutions into guidelines that can improve future generation~\citep{chen2023introspective}, and using multiple LLMs as debating agents to refine feedback for a solution~\citep{du2023improving, liang2023encouraging}. However, they are all evaluated on different datasets and whether the baselines are computed or cost-matched is rarely considered. Notable exceptions are~\citet{shinn2023reflexion} where they consider performance as a function of the number of queries to the language model and~\citet{olausson2023demystifying} which evaluates the performance of a self-debug strategy for code generation as a function of the number of tokens generated.

\subsection{LLM output evaluation}

There has been considerable work on evaluating the output of LLMs -- both via training custom models as well as using the LLMs themselves for self-evaluation. For trained verifiers/rerankers, in~\citet{cobbe2021training}, they train a verifier to rerank outputs of language models for math word problems and show strong improvements. In~\citet{inala2022fault}, they do the same except for code generation. In~\citet{uesato2022solving,yang2022generating}, they train an evaluator for each step in a chain-of-thought and rerank using the combined score for each step in the chain. In~\citet{li2023making}, they weight the self-consistency by the trained verifier confidence. There has also been work recently on using the LLMs themselves to evaluate their own generations. In~\citet{bai2022constitutional}, they use LLMs to do pairwise comparisons between generations achieving high accuracy. In~\citet{ling2023deductive}, self-consistency for every step is used to evaluate how correct a deductive step is. While they can obtain high accuracy as to whether a step is valid or not, they are unable to improve the overall accuracy of answer generation using that.~\citet{tian2023just}
examine multiple strategies for eliciting LLM self-evaluation that is as calibrated as possible. The self-refine~\citep{madaan2023self} approach uses LLMs to get detailed self-evaluation which is used to improve the next round of generation. The Tree-of-Thoughts~\citep{yao2023tree} paper uses LLM self-evaluation to rank which node to explore next.

\section{Inference Budget of Reasoning Strategies} \label{sec:inference_budget}

While the raw performance of different prompting or reasoning strategies for LLMs is a common topic, how different strategies perform when \textit{budget-aware} is less well-studied (with the notable exception of ~\citet{olausson2023demystifying}). However, taking budget into account can be critical when using LLMs. In this section, we describe different usage scenarios that a user could be interested in and what budgetary metrics would be relevant to those scenarios. We furthermore describe how different reasoning strategies can scale in terms of each budget.

\subsection{Budget} \label{sec:budget}

We examine various budgetary metrics for LLMs. Given that the number of input and output tokens often feature prominently across these metrics, we designate them as \(n_I\) and \(n_O\) respectively. 
%\ben{tie back to introduction -- mention that budget is in abstract sense. latency can be considered as one of the budget.}

\begin{enumerate}
\item \textbf{API monetary cost} is generally represented as \(c = \alpha_1 \cdot n_I + \alpha_2 \cdot n_O\). Here, \(n_I\) and \(n_O\) correspond to the number of input and output tokens. The coefficients \(\alpha_1\) and \(\alpha_2\) are specific to the LLM API in use. It's worth noting that in scenarios involving parallel sampling of multiple outputs with a singular input, \(n_I\) is counted once, whereas \(n_O\) is tallied based on the sample count.
\item \textbf{Total number of tokens}, a straightforward metric, is described by \(t = n_I + n_O\). This becomes pertinent when \(\alpha_1 = \alpha_2\), which is true for many LLM APIs and is also reflective of the compute cost. Its simplicity ensures it doesn't inherently favor any specific model or API provider. %\ben{mention that this is directly related to latency}
\item \textbf{Number of queries} of planned API calls can be a rough proxy for the budget. Such numbers can be determined before inference, which can give us rough guidance before performing each reasoning strategy. 
Note that in case we want to sample multiple outputs from the LLM, we count those as \textit{separate} queries

\end{enumerate}
% to include in FAQ
% Other types of budget we can also consider is latency. For instance, budgets that entail the same amount total tokens or monetary cost may have different latency profile. Some reasoning strategies can be more parallelized such as self-consistency over parallel samples, versus some are more serial in nature.

\section{A Critical Evaluation in Budget-Aware Environments}

This section aims to explore key components that can make reasoning strategies successful from the scale-aware perspective. First, we show that the inference budget is often overlooked but is one of the primary indicators of the success of a reasoning strategy. We show that the scale-aware evaluation perspective, CoT (or variants of it like Plan and Solve, Least to Most) self-consistency, for instance, is a strong baseline that can outperform or match many proposed reasoning strategies in the literature given the same level of budget.

% We claim that using a simple Self-Consistency baseline will, in many cases, outperform more complex reasoning strategies proposed in the literature. \ben{this is somewhat known}

We use existing reasoning strategies in literature to perform this study, namely Multi-Agent Debate (MAD) \citep{liang2023encouraging}, Reflexion \citep{shinn2023reflexion}, Plan and Solve \cite{wang2023planandsolve}, Least to Most Prompting \cite{zhou2022leasttomost}, Progressive Hint Prompting \cite{zheng2023progressivehint}, and Tree-of-Thoughts~\citep{yao2023tree}. We conducted our experiments across a diverse range of reasoning tasks, utilizing math reasoning datasets such as GSM8k \citep{cobbe2021training}, MATH \citep{hendrycks2021measuring}, and  TheoremQA\citep{chen2023theoremqa}, along with the commonsense reasoning task CSQA \citep{talmor-etal-2019-commonsenseqa}, and the multi-hop reasoning task HotpotQA \citep{yang-etal-2018-hotpotqa} (see Appendix \ref{appendix:models_datasets}). Additionally, we performed an in-depth analysis of the puzzle game Game24 \citep{yao2023tree} to further our investigation on budget-aware evaluation. We use GPT-3.5 and GPT-4 for our experiments. (See Appendix \ref{appendix:models_datasets} for more details about model hyperparameters)

%In Section \ref{sec:results:MAD_Reflexion} we will demonstrate the evaluation results for MAD and Reflexion and show that CoT SC outperforms them using same or less budget. Then, in Section \ref{sec:results:three_other} we will present the results on three other reasoning strategies. Finally, we will demonstrate how a more complex reasoning strategy (Tree-of-Thoughts) is only competitive when using a stronger model in Section \ref{sec:results:ToT}.

\begin{figure*}[ht!]
\begin{center}
% \vspace{-0.1cm}
% \begin{subfigure}[b]{\textwidth}
% \centering
% % \includegraphics[width=0.85\textwidth]{figures/MAD_Reflexion/GPT-3.5-Turbo-0301_Queries.pdf}
% \caption{GPT-3.5-Turbo-0301: Performance@Number of Queries} \label{fig:number_of_queries}
% \end{subfigure}
% \begin{subfigure}[b]{\textwidth}
% \centering
% % \includegraphics[width=0.85\textwidth]{figures/MAD_Reflexion/GPT-3.5-Turbo-0301_Tokens.pdf}
% \caption{GPT-3.5-Turbo-0301: Performance@Number of Tokens} \label{fig:total_tokens}
% \end{subfigure}
\includegraphics[width=1\textwidth]{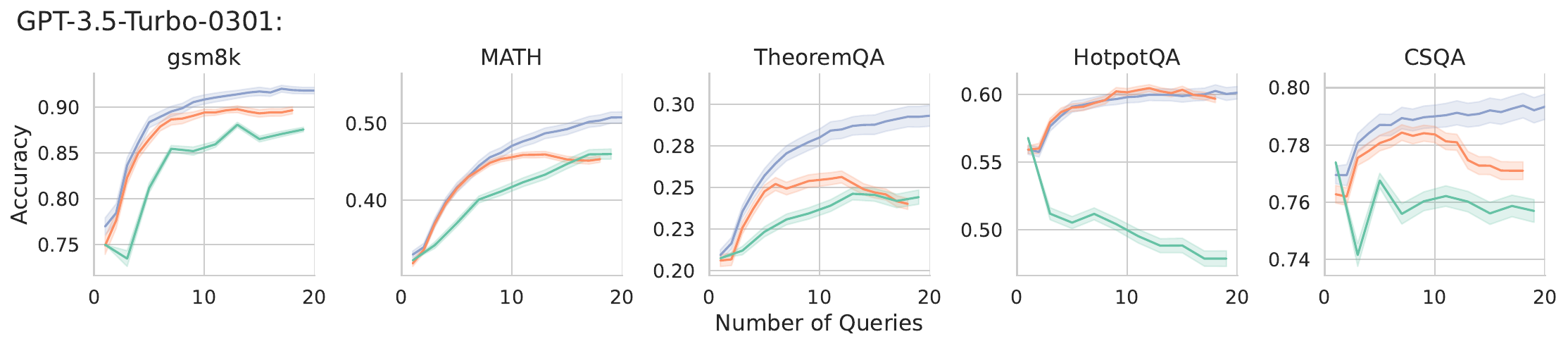}
\includegraphics[width=1\textwidth]{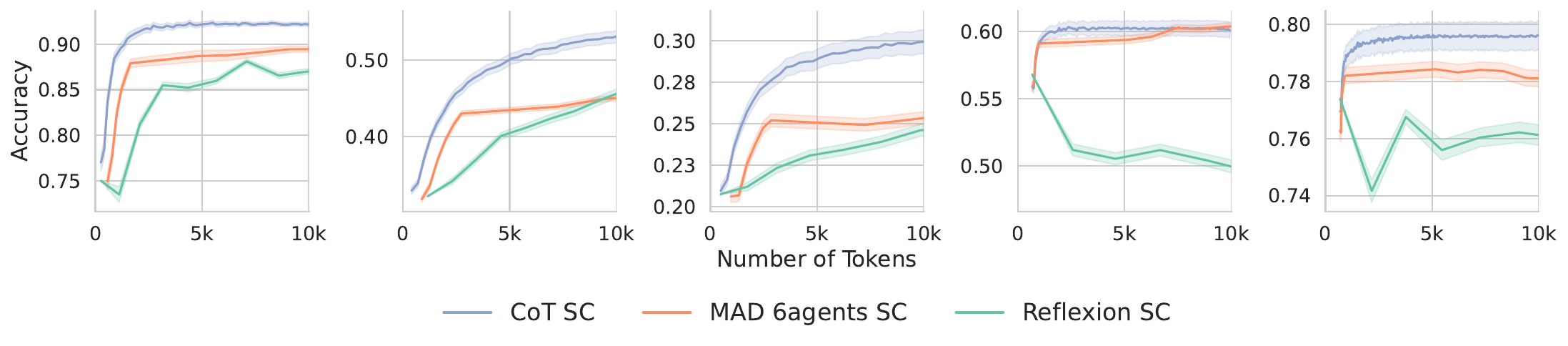}
\includegraphics[width=1\textwidth]{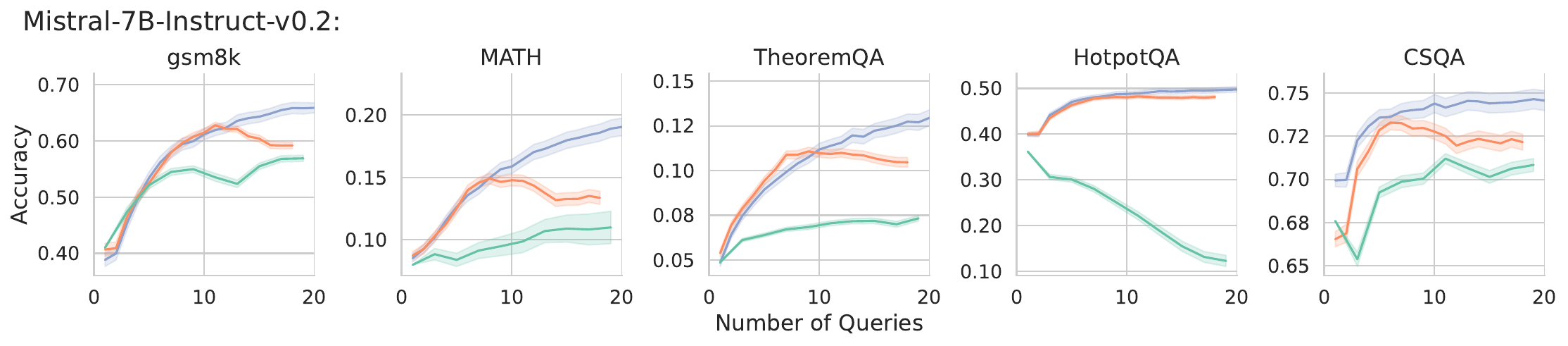}
\includegraphics[width=1\textwidth]{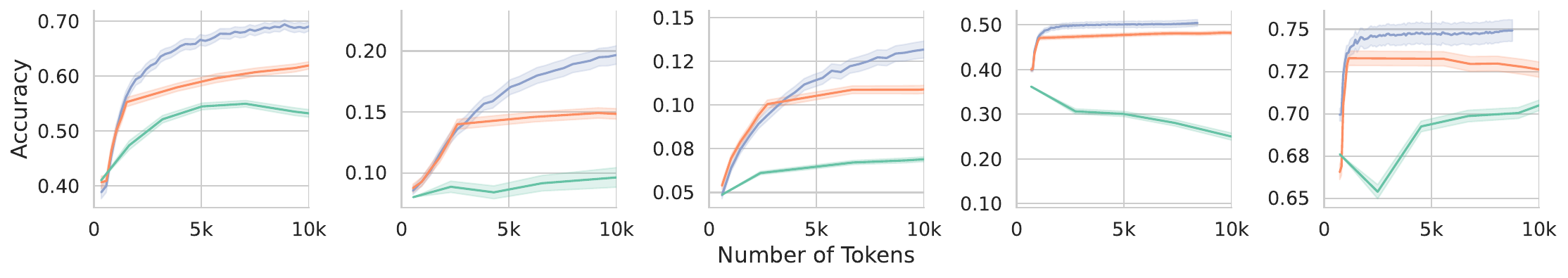}
\end{center}
\vspace*{-3mm}
\caption{Performance@Number of Queries and Performance@Number of Tokens Plots for all 5 datasets. All three methods CoT SC, MAD, and Reflexion are plotted on two models (more models in Appendix \ref{appendix:more_models}). All experiments here are run until at least 10k tokens. CoT with SC consistently beat other reasoning strategies across all 5 datasets with significantly less budget. The budget difference is even more drastic when counting the number of tokens. The MAD result is shown non-round-wise. } \label{fig:budget_queries_vs_tokens}
\end{figure*}

\begin{figure*}[h]
\begin{center}
\centering
\includegraphics[width=1\textwidth]{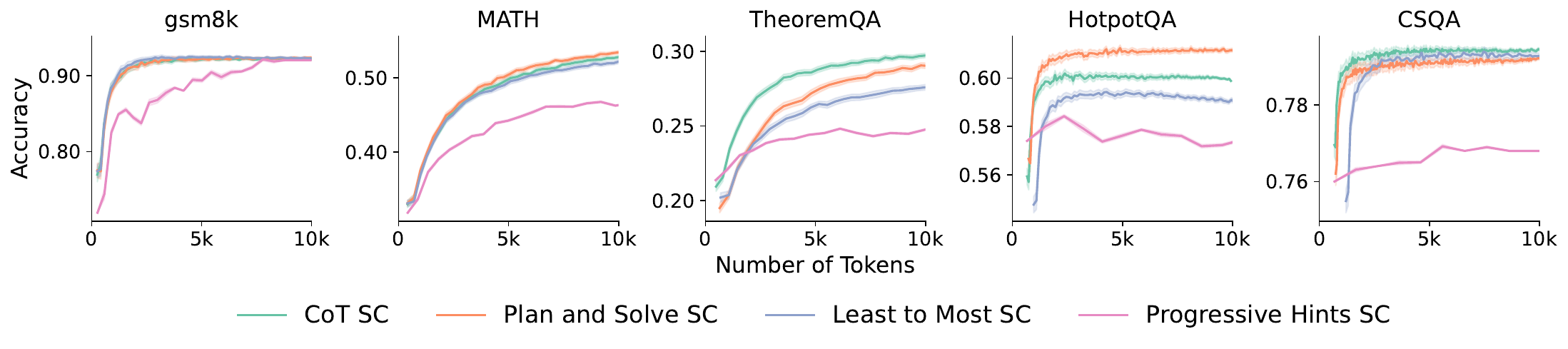}
\end{center}
\vspace*{-3mm}
\caption{GPT-3.5-0301 Performance@Number of Tokens for all 5 datasets using three other strategies: Plan and Solve, Least to Most, Progressive Hints. } \label{fig:0301_new_budget_queries_vs_tokens2}
\end{figure*}

% \subsection{Importance of inference scale} \label{sec:importance_scale}
%  Our findings indicate that the apparent success of novel reasoning methods primarily stems from the influence of scale. 

\subsection{Inference budget unveils superiority of self-consistency baseline over MAD \& Reflexion}
\label{sec:results:MAD_Reflexion}

We present that the observed improvements in performance for various reasoning methods may be strongly influenced by the use of a higher inference budget, rather than the intrinsic merit of the techniques themselves.

Results in Figure \ref{fig:inference_scale} and \ref{fig:budget_queries_vs_tokens} elucidate the efficacy of reasoning techniques, including MAD and Reflexion, in contrast with the SC baseline.
 We keep the budget for each question to be a maximum of 20 queries or 10k tokens. This means for CoT SC, we would sample 20 times. For MAD, we set the number of agents to 6 and the number of rounds to 3 which resulted in exactly 18 queries. For Reflxion, we set it to reflect a maximum of 10 trials (10 proposals 9 reflections for a total of 19 queries). We demonstrate two budgetary metrics which were discussed in (Section \ref{sec:budget}): a) Performance@Number of Queries b) Performance@Number of Tokens. As illustrated in Figure \ref{fig:inference_scale} and \ref{fig:budget_queries_vs_tokens}, when the inference budget of the baseline is aligned with that of each reasoning approach, the perceived benefits of the innovative strategies no longer apply. The SC baseline regularly outperforms more complex strategies when given equivalent budgets across all datasets except HotpotQA. However, even for HotpotQA, the SC baseline still performs as well as other reasoning strategies. Reflexion consistently performs the worst out of the three strategies analyzed and we will discuss this more in detail later. Relying solely on scale-independent assessments, as is sometimes done in prior works, might lead to incomplete or potentially misleading interpretations.

\subsection{Plan and Solve, Least to Most, Progressive Hints performance gain primarily from increased budget}\label{sec:results:three_other}

In this study, we assess the efficacy of the proposed budget-aware metrics across three additional reasoning strategies: Plan and Solve, Least to Most, and Progressive Hints, as depicted in Figure \ref{fig:0301_new_budget_queries_vs_tokens2}. Among these strategies, chain-of-thought self-consistency (CoT SC) emerges as a competitive approach, outperforming others in most scenarios except in the HotpotQA dataset. Here, Plan and Solve, when coupled with Self-Consistency (SC), surpasses CoT SC. It's pertinent to recognize Plan and Solve and Least to Most as specialized iterations of the CoT approach. Specifically, Plan and Solve directs the LMs to strategize prior to resolving the query, whereas Least to Most deconstructs the question before answering. Thus, both Plan and Solve SC and Least to Most SC can be conceptualized as nuanced versions of CoT SC. Conversely, Progressive Hints, which leverages sequential answers as cues for subsequent questions, exhibits the least effective performance. This comparative analysis underscores that the observed improvements in performance are primarily attributable to increased budget allocations rather than the inherent advantages of the methodologies. Some complex reasoning strategies perform seemly perform better, however, when accounting for budget, fail to beat basic CoT with self-conssitency.
Evaluations on more datasets and types of budget as well as details about each strategy can be found in Appendix \ref{appendix:more_models:three_other_strategies}.

\subsection{Tree-of-Thoughts is competitive with a caveat}\label{sec:results:ToT}

\textbf{Tree-of-Thoughts can outperform baselines} We in addition evaluated the well-known Tree-of-Thoughts strategy in a scale-aware manner on the logical game Game of 24. Notable discrepancies emerged in the behavior of the model when transitioning to GPT-4 and we modified our budget-aware metric slightly to further scrutinize GPT-4. 
%  Our findings mirrored the above conclusions when GPT-3.5 is used, where the integration of CoT coupled with Self-Consistency exhibited improved performance with reduced computational resources. However, n

\textbf{A strong model is needed to perform better than baseline} In Figure \ref{fig:CoT_vs_ToT}, we show the performance of GPT-3.5 with the Tree-of-Thoughts reasoning strategy on Game of 24\footnote{We used a modified thought evaluation prompt for GPT-3.5 that gave much better results than the default one}.  The performance of Tree-of-Thoughts lags that of a simple SC by a considerable margin. This is in stark contrast to the GPT-4 results with Tree-of-thoughts where all other strategies plateau very early and Tree-of-Thoughts beats all of them by a big margin. This remains the case even when we account for the budget (query or token budget) as the other strategies have a low performance ceiling. However, note that Tree-of-Thoughts requires a significant budget commitment to deliver such a performance. On weaker models than GPT-4, it is still better to use simpler strategies like SC which outperforms ToT by a considerable margin (Figure \ref{fig:CoT_vs_ToT}).

\begin{figure}[t!]
\begin{center}
\includegraphics[width=0.48\textwidth]{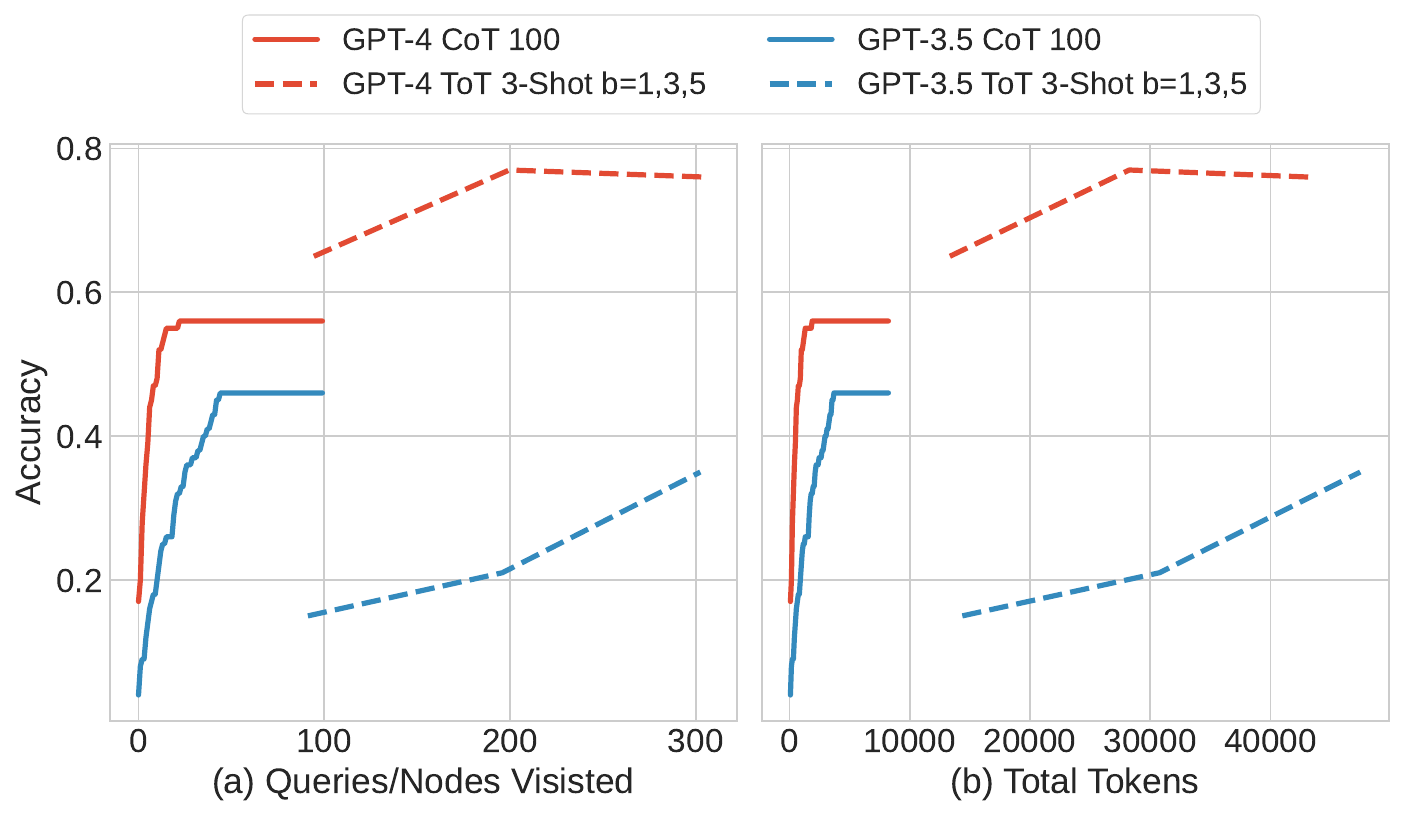}
% \begin{subfigure}[b]{0.24\textwidth}
% \includegraphics[width=1\textwidth]{figures/figures_game24/queries_CoT_vs_ToT_allnodes.pdf}
% \caption{Queries/Nodes Visited}
% \end{subfigure}
% \hspace*{-2mm}
% \begin{subfigure}[b]{0.24\textwidth}
% \includegraphics[width=1\textwidth]{figures/figures_game24/tokens_CoT_vs_ToT.pdf}
% \caption{Total Tokens}
% \end{subfigure}
% \hspace*{-1mm}
\end{center}
\caption{ToT vs. CoT SC for both GPT-3.5 and GPT-4 on Game of 24. The dotted lines represent the performance of ToT. For ToT using GPT-4 (red), results for three settings are included. All CoT results are computed using self-consistency on 100 samples.
}
\label{fig:CoT_vs_ToT}
\end{figure}

%\subsection{Does a higher inference budget always lead to better reasoning?}

\section{What Makes Reasoning Strategies Work } \label{sec:Analysis}
We conduct a more in-depth analysis into reasoning strategies. Specifically, we investigate what causes the performance gap evidenced by our budget-aware evaluations among reasoning strategies like MAD or Reflexion versus self-consistency in Section \ref{sec:reasoning_not_equal}. In Section \ref{sec:ToT_ablation}, we delve deeper into different components of tree-of-thoughts. Finally in Section \ref{sec:self_evaluation} we examine the effectiveness of self-evaluation in the reasoning loop. 

\subsection{Reasoning strategies do not benefit equally from higher inference budget}\label{sec:reasoning_not_equal}

The scale-aware perspective offers clear a guideline for what reasoning strategies make sense. 
That is, a proposed reasoning strategy should be considered effective only if its performance is better \emph{compared to a baseline of equivalent budget}; otherwise, the incurred cost could not be justified since the baseline with comparable cost performs better (in terms of FLOPs, latency, monetary cost, latency, or any other type of budget we may care about). A question arises whether we can keep increasing the budget to obtain the best possible abilities.

As seen in Figure \ref{fig:budget_queries_vs_tokens}, we find that the self-consistency baseline exhibits a smooth increase in scores with respect to scale. However, such a trend does not always hold for other reasoning strategies. 
For instance, in multi-agent debate, an augmented inference budget eventually experiences a performance plateau (potentially due to diversity decrease which we analyze later). For the MAD setting with 6 agents, the graph for MAD and SC overlaps up to 6 queries since they correspond to the exact same strategy up to that point. After 6 queries, the MAD strategy switches to the second round where the performance gain noticeably lessens compared to self-consistency. The amount of tokens required for each subsequent round also increases drastically since each agent needs to look at rounds' conversations.
The lowered performance compared to the SC baseline may arise because subsequent rounds of MAD may incite a cascading effect of cumulative mistakes, or snowballed hallucinations suggested in \citet{hallucinations_snowball}. Below, we offer some further explanations behind the performance gap.

\begin{figure}[t!]
% \vspace{-1mm}
\begin{center}
%\framebox[4.0in]{$\;$}
\includegraphics[width=0.43\textwidth]{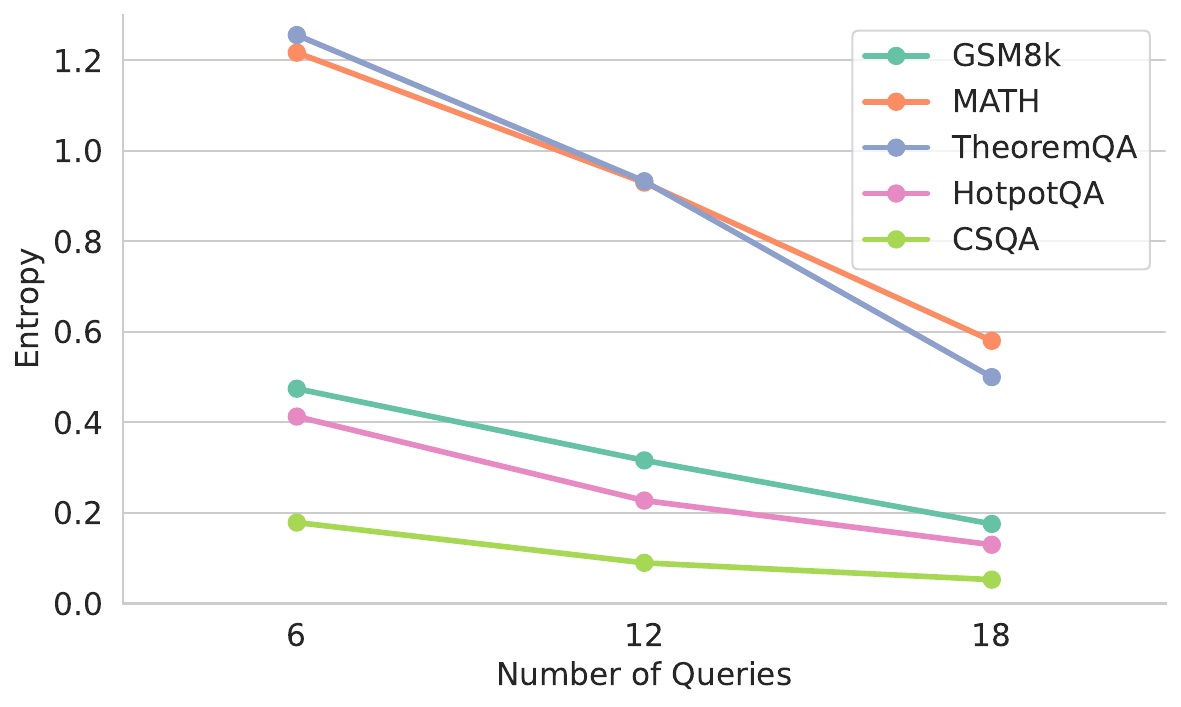}
\end{center}
\caption{The diversity measure of the answers proposed by GPT-3.5. The strategy is MAD with 6 agents and 3 rounds.
} 
\label{fig:diversity}
% \vspace{-2mm}
\end{figure}

\subsubsection{Dependent sampling can hurt response diversity}
% \ben{not necessarily the complexity but more on the chained reasoning / hallucination.
% for instance, if a strategy is 'complex' but involves independent sampling, it may not suffer the same way.
% Can we be more precise about this?
% }\junlin{make the description more accurate}

Multi-agent debate conditions on the previous round's answers to sample new answers. We posit one of the reasons multi-agent debate performs worse as the budget increases is because it reduces response diversity (through this dependent sampling) and, hence is more likely to tunnel on the wrong answer. To show this, we compared the entropy of the solutions generated at each round for MAD vs. SC. The results are in Figure \ref{fig:diversity}. Here, the entropy corresponds to the diversity of answers within each round. The entropy consistently declines for multi-agent debate as we move to the next round suggesting exactly the kind of cascading effect we hypothesized. By contrast, self-consistency does not suffer such negative consequences and even increases its solution diversity since the responses are generated independently without affecting one another.

%In this section, we explores the behavior of complex reasoning strategies such as MAD and the self-consistency baseline with respect to scale. 

\begin{figure}[t!]
% \vspace{-1mm}
\begin{center}
\includegraphics[width=0.43\textwidth]{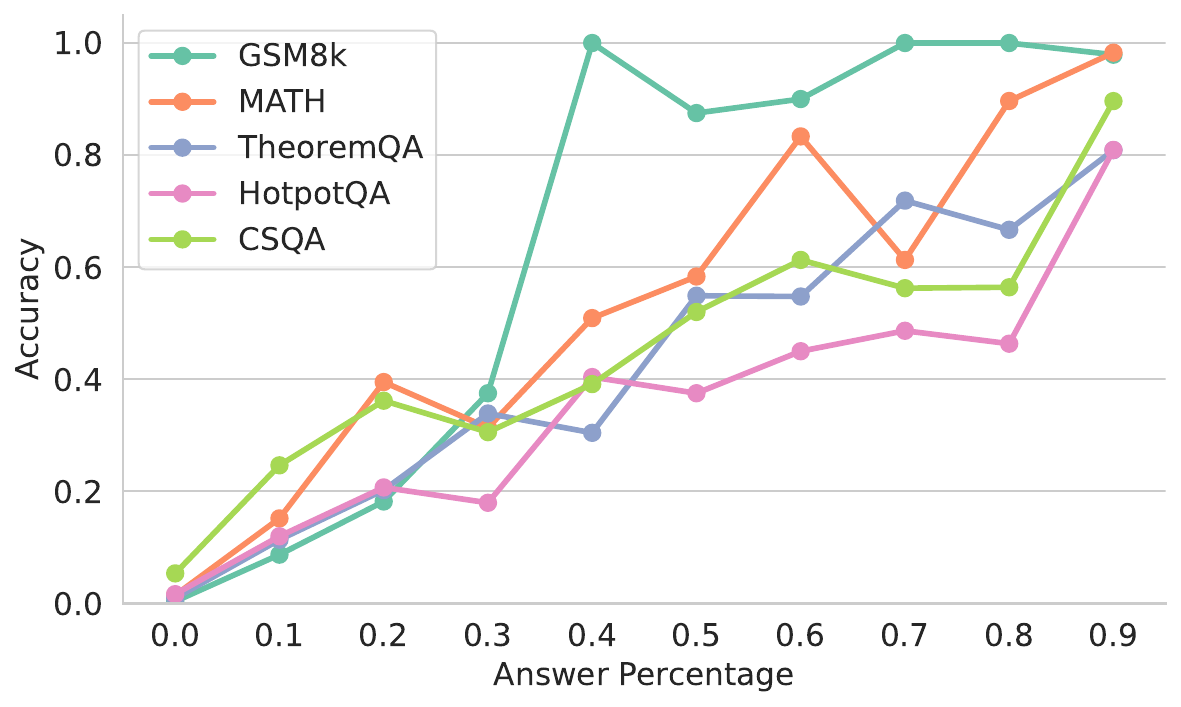}
\end{center}
\caption{Calibration result binned by answer percentages. If an answer appears more times within all the samples, that answer is more likely to be the correct answer.}\label{fig:calibration_answer_popularity}
% \vspace{-8mm}
\end{figure}

\subsubsection{Effectiveness of independent sampling with chain-of-thought prompting} \label{sec:4.2.2}

Next, we outline a framework that helps explain what makes self-consistency successful. We first empirically verified that the higher the occurrence of an answer, the more likely it is the correct answer (Figure \ref{fig:calibration_answer_popularity}). Self-consistency is able to capitalize on this and hence improves performance as the budget increases.

We model the answer generation process by language models as a binomial distribution where each problem has an inherent probability $p_i$ of being answered correctly.  This analysis reveals several insights:
% The analysis provides specific insights and observations that are crucial for understanding the improvement in a model's performance as the number of trials increases.

\begin{figure*}[t!]
% \vspace{-0.1cm}
\begin{center}
\includegraphics[width=0.95\textwidth]{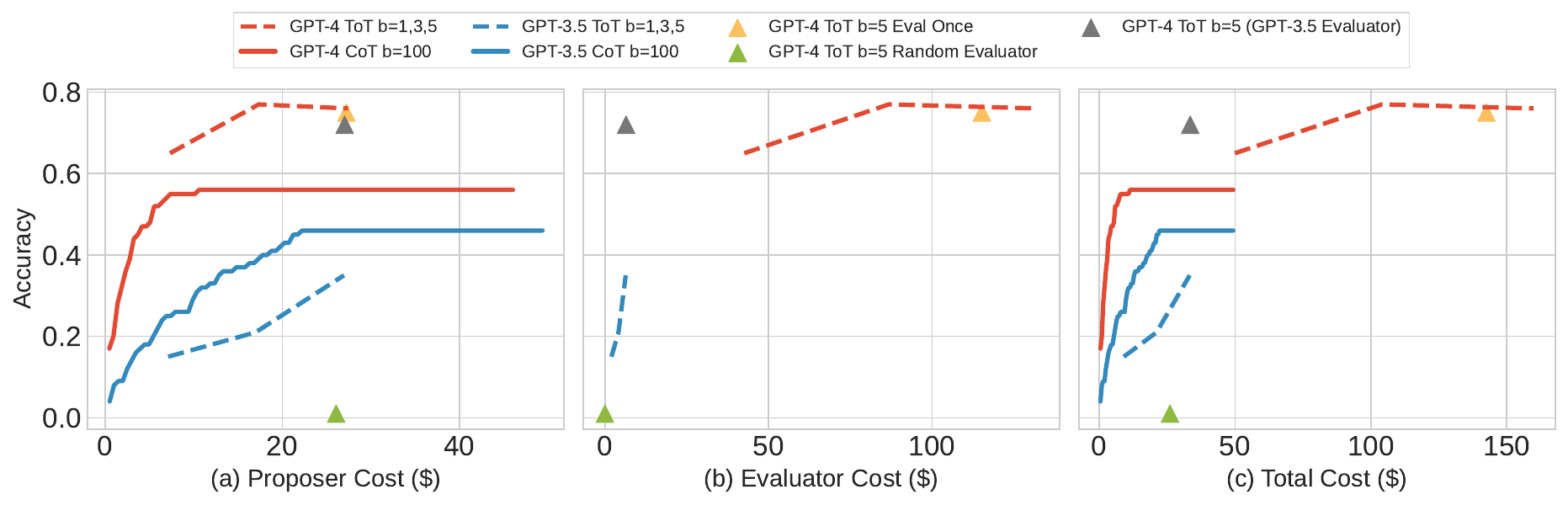}
\end{center}
  \caption{Thought proposer and and thought evaluator budget on the dataset of Game of 24. For the cost computation, we assumed the prices OpenAI had as of Aug 14, 2023 which were \$0.002 per 1K tokens for GPT-3.5 (encoding or decoding), \$0.03 per 1K encoded tokens and \$0.06 per 1K decoded tokens for GPT-4.}
  \label{fig:accuracy_vs_budget_ToT_Game24}
\end{figure*}

\begin{figure}[h]
% \vspace{-2mm}
\begin{center}
\includegraphics[width=0.43\textwidth]{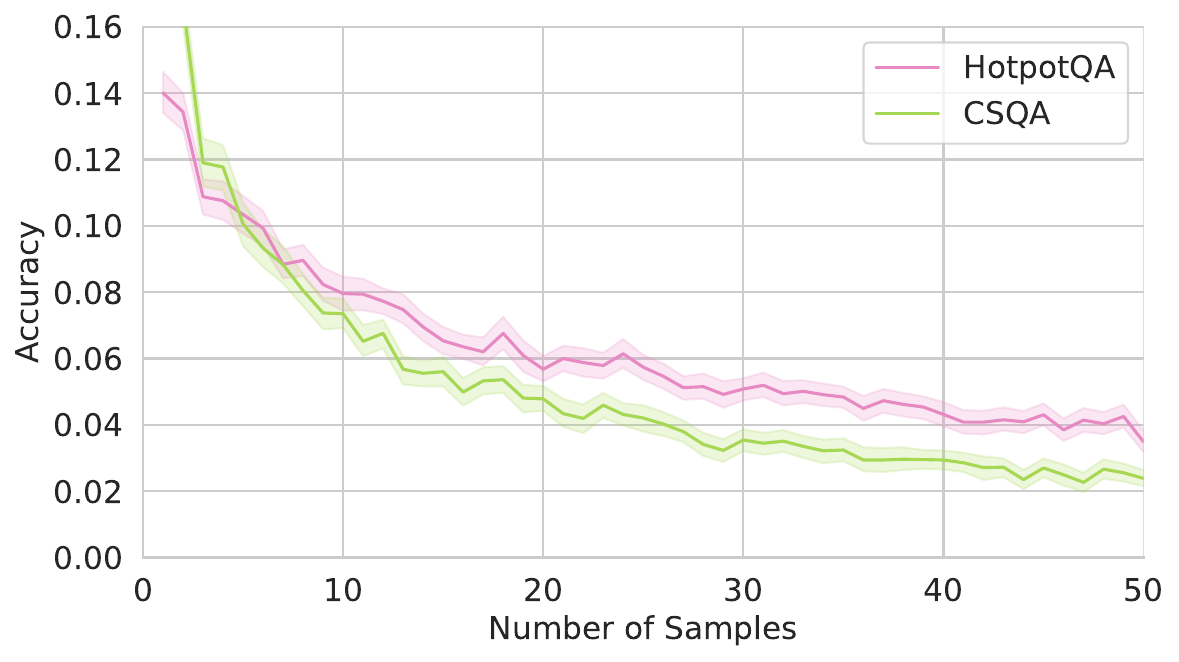}
% \hspace*{-6mm}
% \vspace{-2mm}
\end{center}
\caption{SC making things worse on QA problems. We selected a subset of problems where the correct answer is not the majority. For this subset, the performance decreases with more samples. This is empirical evidence that SC may cause worse performance if the individual sample's accuracy is low.}
\label{fig:SC_Worse}
\end{figure}

\begin{enumerate}
  \item \textbf{Convergence}: The probability of a correct majority vote converges to 0 or 1 as the number of trials increases, depending on whether the probability of a correct answer \( p_i \) is less or greater than 0.5.

  \item \textbf{Speed of Convergence}: Convergence is fast for extreme values of \( p_i \) (closer to 1 or 0), but slow if \( p_i \) is near 0.5.

  \item \textbf{Distribution of Correctness}: By placing a prior on \( p_i \) (for instance, with a beta distribution), the aggregate score over the entire dataset converge to non-extreme values,  resembling the behavior observed in our results.

\end{enumerate}

  %Our findings indicate that the model's performance improves with the number of trials if the distribution of \( p_i \) has a mean greater than 0.5. If the mean is lower than 0.5, then performance can degrade over time (empirically shown in Figure \ref{fig:SC_Worse}).
  That is, self-consistency performance increases smoothly over time is due to the artifact of a model consistently answering plausible answers that tend to be more correct than not. In Appendix \ref{sec:consistency_math}, we detail the analysis with extension to a multinomial setting with Dirichlet priors.  
  %The rate of convergence and the effectiveness of majority voting depend on the value of \( p_i \), and this framework can be harnessed to enhance reasoning tasks.
  % \vspace{-2mm}

% \vspace{-2mm}

% \sid{Describe binomial and multinomial formuation. Give intuition as to why convergence behavior happens and give plots. Reference tail bounds in appendix}
% \sid{@junlin Include plots of SC making things worse for the 97 problem subset and for other datasets}
% %\paragraph{Performance of majority voting consistently rises with increased inference budget.} 
% \paragraph{Majority vote selects a correct example \ben{ECE results} }
% % say something along the line of calibration
% % model seems to have ok calibration
% % and the strategy is able to select a correct example
% \ben{ relate this to expected calibration error}
% \paragraph{A correct sample may be found given scale}
% In Figure \ref{fig:oracle_nonoracle}, we demonstrate the performance of the oracle setting where the oracle can indicate whether a correct answer is generated and selects it as a final answer. This setting serves as an upper bound of the performance of each reasoning strategy. We find that the oracle performance for all reasoning strategies increases, but notably more rapidly in parallel sampling alone. This is especially the case for parallel sampling where, as shown in Figure X, the diversity of the generated answers rapidly increases whereas it declines for strategies like MAD.

% \vspace{-2mm}

\begin{figure}[h!]
\begin{center}
\centering
\includegraphics[width=0.43\textwidth]{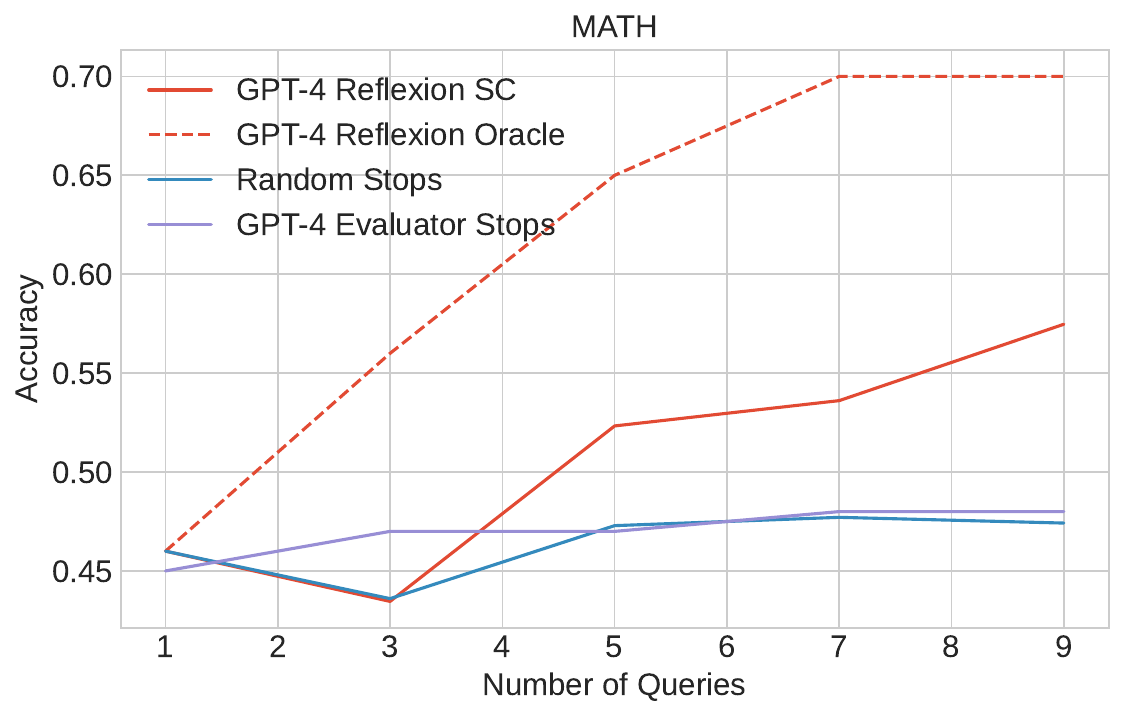}
\caption{Ablation Study of the effect of evaluator on Reflexion with GPT-4. Having an oracle evaluator outperforms SC.}
\label{fig:reflexion_evaluator}
\end{center}

\label{fig:evaluator}
\end{figure}

\subsection{Tree-of-thoughts}\label{sec:ToT_ablation}

% Several of the strategies presented here have a self-evaluation component which can eat up a lot of the budget. For example Tree-of-Thoughts has to make 1 query for every candidate thought to evaluate its goodness. To assess we need to allocate an equal budget to the evaluation component as we do to the answer generation component, we first examine how strongly the performance of various strategies depends on the self-evaluation ability of the model. Reflexion relies on a perfect evaluator to decide whether to keep reflecting or stop and return the answer (in the non-oracle mode). Multi-Agent Debate relies on self-evaluation in a subtler way -- the LLM has to implicitly evaluate the answers generated in the last round and use that assessment to generate the next round of answers. Tree-of-Thoughts does self-evaluation at every node in the tree to decide which thoughts to keep and which to prune. In Figure \ref{fig:evaluator}, we assess an evaluation strategy that is no better than random (deciding to stop randomly after a certain number of reflection steps), using the LLM to decide whether the most recent answer is correct and stopping based on that or using a perfect "oracle" evaluator. Random selection of the stopping point gives poor results. However when augmented with the oracle evaluator, there is a big jump in performance. 

In this section, we investigate the factors that contribute to the enhanced performance of the tree-of-thoughts strategy when compared to the self-consistency baseline. Tree-of-thoughts strategy mainly has two components: a proposer and a self-evaluator. The proposer proposes intermediate steps or answers and the evaluator decides whether to prune or continue on current branches. Hence we further divide the budget into the \textbf{proposer budget} and the \textbf{evaluator budget}. We aim to answer questions like how much of the performance can be attributed to self-evaluation ability.

For the ablation study, we compare four setups for tree-of-thoughts on the Game of 24.
\begin{enumerate}
    \item The standard tree-of-thoughts strategy where we use GPT-4 to evaluate the new thoughts.
    \item The standard tree-of-thoughts strategy except we now do an evaluation of a thought only once as opposed to 3 times.
    \item Using a weaker model (GPT-3.5) as the evaluator while using GPT-4 as the thought generator.
    \item Random evaluator, where we randomly select the subset of thoughts to prune.
\end{enumerate}

\begin{figure*}[t!]
\begin{center}
\includegraphics[width=1\textwidth]{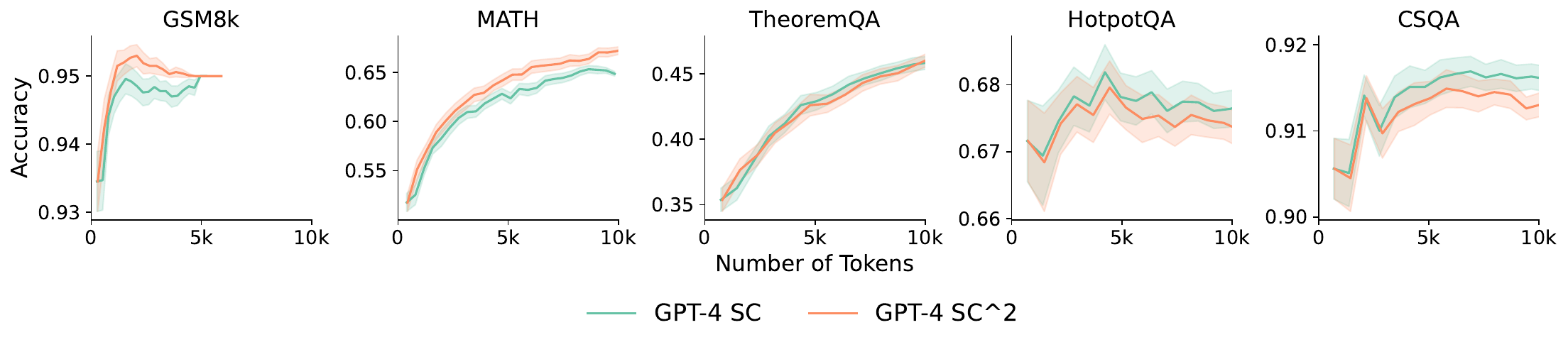}
\end{center}
\caption{$SC^2$ with total tokens being the budget if caching is enabled.
} \label{fig:tot} 
\label{fig:WCSC_All_datasets_cached}
\end{figure*}

\paragraph{Evaluator quality has a non-trivial impact}
As observed in Figure \ref{fig:accuracy_vs_budget_ToT_Game24}, a random evaluator leads to a very steep performance drop for ToT for both best@k as well as total accuracy. Results imply that an evaluator has a non-trivial impact. Evaluation is done only once per thought as opposed to multiple times also leads to significant performance drops. 

\paragraph{Cost-efficiency of evaluator}
However, if we use a weaker evaluator like GPT-3.5, we can maintain most of the performance while being very cost-efficient. For example, running Tree-of-Thoughts (with beam width set to 5) with GPT-4 as thought proposer and GPT-3.5 as evaluator on 100 Game of 24 instances costs \$33.53 while getting an accuracy of 72\%. Using GPT-4 as the evaluator on the other hand increases the cost almost 5x to \$159.87 while only improving accuracy to 76\%. Even if we restrict ourselves to $b=1$ with GPT-4 as an evaluator, we still get a higher cost of \$49.9 while getting a worse performance of 65\%. 

\paragraph{More effective use of budget for proposer}
We found that the proposer plays a significant role in performance, in addition to the evaluator. For further ablation results, please refer to Appendix Table \ref{table:ToT_performances}. Using a GPT-3.5 proposer with a GPT-4 evaluator yielded an accuracy of only 0.38, which is considerably lower than when using GPT-4 for both roles. We do not delve deeply into this topic here, as most reasoning strategies require some form of proposer. Our primary focus lies in assessing the extent of the advantages conferred by the unique self-evaluator component, though we wanted to highlight the importance of the proposer in this context.

\subsection{Reflexion ablation study}
 We conducted a similar experiment on Reflexion shown in Figure \ref{fig:reflexion_evaluator}. We compared standard Reflexion with oracle where an oracle indicates the correct answer, Reflexion with self-consistency, Reflexion with random evaluator, and Reflexion with GPT-4 as evaluator. We found that although Reflexion with oracle improves performance over SC by a large margin, Reflexion with GPT-4 Evaluator still underperforms Reflexion with SC. This highlights the substantial gap between a perfect evaluator and a GPT-4 evaluator for this dataset, indicating that the self-evaluation capability of LLMs has considerable room for improvement. In addition, evidence suggests that doing Reflection is only slightly better than random guessing.  LLMs as a better evaluator could unlock significantly more performance.
 
 % From last section we knew that self-evaluation is model dependent, we additionally posit that self-evaluation strategy and dataset dependent. \ben{need to unpack a lot more}

\subsection{Self-evaluation is a promising budget-efficient improvement but is currently lacking}
\label{sec:self_evaluation}

% \ben{in a way this is not very satisfying}
Self-evaluation usually involves very few tokens generated since evaluation is short. This can be potentially very cost-effective since prefilling is cheaper and faster.
In this section, we investigate further how a self-evaluator can benefit the reasoning process in a budget-aware setting and demonstrate why there may still be a long way to go.
In this section, we investigate further how a self-evaluator can benefit the reasoning process in a budget-aware setting. 

\paragraph{Accuracy and Calibration of Self-Evaluation}

% In Table \ref{table:self_eval_accuracy_MATH}, we show the accuracy of self-evaluation of GPT-4 on the MATH dataset as a function of the number of samples. The binary strategy performs the best by far in terms of both accuracy and calibration. The numerical confidence strategy is worse with the confidence probability strategy being the worst by far. 
% In Figure \ref{fig:self_eval_calibration_MATH} we show the calibration plots. Ideally, we want self-evaluation to be well calibrated, that is, higher confidence should imply higher accuracy and vice versa. We see that this is the case for easier datasets like GSM8K, but is not the case for harder datasets like MATH.

To arrive at the final format of doing self-evaluation, we benchmarked three types of evaluations: Yes or No, Score 1-10, and Probability between 0 to 1. Through our benchmark we found Yes or No to be the most calibrated. More details and calibration results can be found in Appendix \ref{sec: self-evaluation}

\begin{table}[ht]
\centering
\resizebox{\columnwidth}{!}{%
\begin{tabular}{ c c c c } 
 \toprule
 Dataset & Correct Accuracy & Incorrect Accuracy & Total Accuracy \\ [0.5ex] 
 \midrule
 GSM8K & 0.992 & 0.156 & 0.937 \\ 
 MATH & 0.911 & 0.461 & 0.707 \\
 TheoremQA & 0.945 & 0.232 & 0.547 \\
 HotpotQA & 0.994 & 0.029 & 0.675 \\
 CSQA & 0.987 & 0.06 & 0.901 \\
 \bottomrule
\end{tabular}
}
\caption{Self-evaluation accuracy on 5 datasets. Correct accuracy denotes self-evaluation accuracy for answers that turn out to be correct. Incorrect accuracy is the self-evaluation accuracy of incorrect answers. All numbers are done with GPT-4-0613}
\label{table:self_eval_accuracy}
\end{table}

\paragraph{Self-evaluation ability on different datasets}

Table \ref{table:self_eval_accuracy} shows the self-evaluation accuracy for GPT-4 for multiple datasets. The self-evaluation accuracy turns out to be heavily dependent on the dataset. It is possible that for problems that are too hard for the model, it ends up weighing the writing \textit{style} of the answer much more heavily than the correctness of all the intermediate steps when doing the evaluation. We saw empirical evidence in Figure \ref{fig:self_eval_calibration_MATH} where easier datasets are better calibrated and vice versa. On harder tasks like TheoremQA, GPT-4's accuracy is close to random. This means LLMs have a long way to go before they are reliable evaluators. We examine this in more detail in the appendix \ref{sec: self-evaluation}.

\paragraph{Self-Confident Self-Consistency ($SC^2$)}
As an investigation of using self-evaluation to improve reasoning procedure, we propose to weigh the SC by the confidence the model has in its answer, derived from self-evaluation. We call this score the \textit{Self-Confident Self-Consistency} ($SC^2$) score. We showed that $SC^2$ beats self-consistency on GSM8k and MATH while fall behind on the other three as shown in Figure \ref{fig:WCSC_All_datasets_cached}.\footnote{More details can be found at Appendix \ref{sec:self-evaluation:sc2}.} This shows that although theoretically, self-evaluation is promising (shown with oracle results in Figure \ref{fig:reflexion_evaluator}), it is still lacking in practice due to low accuracy.

\section{Conclusion} \label{sec:conclusion}

In this paper, we examined the performance of seven reasoning strategies on the often overlooked metric of budget. We used budget metrics of queries and tokens to reflect various ways LLMs are used (LLM APIs or self-host). We identified self-evaluation as an important aspect of many reasoning strategies and analyzed different prompting strategies to have the model evaluate its generations. We then evaluated self-evaluation and found that although self-evaluation could be promising at improving performance while being cost-effective, current LLMs are mostly incapable of doing that. With the current popularity of reasoning strategies, we think this more balanced budget-aware metric is beneficial for the community and helps set the correct trajectory for future LLM research.

\clearpage
\section{Limitations}

Our goal in the paper was to highlight the importance of different aspects of the generation budget for LLMs that are often ignored in the recent spate of reasoning strategies for LLMs. To that end, we chose some representative reasoning strategies and evaluated them on some common reasoning tasks. However, due to both monetary and time constraints, we could not include even more reasoning strategies or tasks. A more exhaustive evaluation might reveal additional nuances which would be interesting to explore. 

\bibliography{token_economies}
\bibliographystyle{icml2024}

%%%%%%%%%%%%%%%%%%%%%%%%%%%%%%%%%%%%%%%%%%%%%%%%%%%%%%%%%%%%%%%%%%%%%%%%%%%%%%%
%%%%%%%%%%%%%%%%%%%%%%%%%%%%%%%%%%%%%%%%%%%%%%%%%%%%%%%%%%%%%%%%%%%%%%%%%%%%%%%
% APPENDIX
%%%%%%%%%%%%%%%%%%%%%%%%%%%%%%%%%%%%%%%%%%%%%%%%%%%%%%%%%%%%%%%%%%%%%%%%%%%%%%%
%%%%%%%%%%%%%%%%%%%%%%%%%%%%%%%%%%%%%%%%%%%%%%%%%%%%%%%%%%%%%%%%%%%%%%%%%%%%%%%
\newpage
\appendix
% \onecolumn

%\section{Appendix}

% \section*{List of things to get in}

% \begin{itemize}
% \item 
% % \ben{(1) we should emphasize that we propose to use budget-aware evaluation going forward to future development of reasoning strategies. Any reasoning strategy to be adopted should be better against the simple SC baseline. (2) leading to proposing our own strategy -- weighted Majority Vote?}
% \item use implicit confidence based on the frequency of generation to study calibration. contrast this with confidence obtained from asking yes/no.
% \item tie it back to scale for all sections.
% % \item  \ben{ importance of CoT in SC}
% \item another plot we can do it the total tokens without considering repeated context. so total prompt tokens for each round + total generate tokens.
% And we can suggest that this is the budget in the caching scenario where we host LLM. Also this would be the case if API takes an algorithm instead of explicit prompt turn by turn.
% \end{itemize}

\section{Model/Dataset Details}
\subsection{Datasets}
\label{appendix:models_datasets}
Here we describe the datasets we used in our experiments.
\paragraph{GSM8K} GSM8K consists of 8.5K grade school math problems. There are 7.5K examples in the training set and 1K in the testing set. Each problem is expressed in natural language and usually involves multi-hop reasoning.
\paragraph{MATH} MATH dataset collects 12.5K (7.5K training, 5K testing) high-school level competitive math problems in natural languages. This dataset is considerably harder than GSM8K. 
\paragraph{TheoremQA} Theorem QA annotated 800 QA pairs covering over 300 theorems spanning across Math, EE\&CS, Physics and Finance. We focus on math reasoning hence we only used the subset that covers math problems which contains 442 questions. This dataset is even harder than GSM8K since these questions are college-level and involve using theorems.
\paragraph{CSQA} CSQA sourced commonsense reasoning questions from crowd workers based on ConceptNet. It has a total of 12,247 examples (9741, 1140,1140 for the size of train, dev, and test set respectively).
\paragraph{HotpotQA} HotpotQA collects 113K question-answer pairs that require multi-hop reasoning. There are 7,405 pairs in the test set.
\paragraph{Game of 24} Game of 24 is a mathematical reasoning challenge, where the goal is to use 4 numbers and 4 
arithmetic operations (+-*/) to obtain 24. \citep{yao2023tree} collects 100 problems from 4num.com which are ranked 901-1000 (it is ranked from easy to hard, so these 100 are relatively hard). \\

For each dataset above, we randomly sampled 100 samples from the test set for all of our experiments. For Game of 24, since there are exactly 100 problems, we just use the same 100 problems as in \citep{yao2023tree}.
\subsection{Model Hyperparameters}
Since we want to maintain the diversity of reasoning processes, most of the results are obtained with a temperature of 1 for GPT-3.5 and GPT-4. In our preliminary study, we also tested with a temperature of 0.7 and 0.5 and observed the same conclusion. The GPT-3.5 version we used is 0301. The GPT-4 version we used is 0613.

\section{Additional Result for Budget-aware Performance Metrics}

% \subsection{Impact of Different Budget Metrics}
% \ben{need to see if there is any interesting conclusion here. Perhaps we show that it's robust and the number of queries is a good enough representative. }
% \ben{if we can cache -- some reasoning strategies may be more effective wrt budget}

\subsection{Budget Metrics on All Datasets}

Budget metrics on all datasets are shown in Figure \ref{fig:budget_queries_vs_tokens} and Appendix \ref{appendix:more_models}.

\subsection{Detailed description of Reasoning strategies}

\begin{enumerate}
\item \textbf{Tree of thoughts} generates a search tree to search through possible chains of thought. It maintains a chain of thought. At each node in the tree, it generates a list of candidate thoughts to be added to the chain and does an evaluation to select the next thought to add. It concludes by generating an answer at a leaf node of the tree. The path in the tree from the root to the leaf node forms a single chain of thought, with each node corresponding to a single thought. If the answer is deemed incorrect (as per another evaluator), it backtracks to a previous node of the tree (unwinding the chain of thought along the way) and selects the next thought out of the candidate list of thoughts to add to the chain of thought.
\end{enumerate}

% \subsection{Replication with other LLM APIs}
% \ben{we should confirm the conclusion with other LLMs to see if the trend holds}

\subsection{Self-Evaluation with CoT}
All of our self-evaluations are done without CoT. For both evaluation calibration and weighted confidence self-consistency, we only generated one token "yes" or "no" or one number. One may be interested in whether CoT can improve the self-evaluation performance and further boost the results. We tested this by extracting 160 CoT answers from 80 questions from GPT-3.5, where each question we extract 1 correct CoT answer and 1 incorrect CoT answer. We then compared the performance of direct evaluation versus CoT then evaluation. For GPT-3.5-turbo-0301, the accuracy increased from $50.625\%$ to $54.375\%$. For GPT-4-0613, the accuracy increased from $78.75\%$ to $79.375\%$. For GPT-4 the benefit from CoT is very mariginal and we concluded that it is not worth the extract cost from CoT. Hence we use the direct evaluation for all of our self-evaluations.

% \section{Additional Results}
% \subsubsection{Impact of an oracle and "lottery ticket" samples }
% \ben{to continue. }

Figure \ref{fig:reflexion_evaluator} that investigates the Reflexion technique~\citep{shinn2023reflexion} reveals a similar trend compared to the multi-agent debate with respect to inference scale.
We find that Reflexion relies heavily on the oracle that helps the model determine when the correct answer is encountered and stops the generation early and returns that answer. This is in contrast to strategies like SC. We demonstrate the performance of Reflexion including baselines that have access to oracle and without. For direct comparison, it is more fair to compare strategies within the group with access to an oracle, or without. We find that in each group, inference scale is a strong prediction on the performance.

\section{Mathematical Framework for Self-Consistency} \label{sec:consistency_math}

In many real-world reasoning tasks and decision-making processes, the use of SC has emerged as a powerful and often robust technique. Whether it's human experts forming a consensus or ensemble methods in machine learning, the idea of aggregating multiple opinions to reach a final decision has proven to be effective. The empirical success of SC in various domains, such as classification, regression, and human-driven decision-making, motivates a deeper examination into the underlying principles that make it work so well.

For instance, in complex reasoning tasks where individual models or experts might be uncertain, the wisdom of the crowd often leads to improved accuracy. SC can act as a regularization method, mitigating the effects of overfitting or biases that might be present in individual models. By combining multiple models or opinions, SC captures the common patterns among them, enhancing generalization to unseen data.

In this work, we seek to understand what makes SC an effective strategy, especially in the context of reasoning tasks. We aim to analyze the mathematical properties and probabilistic behavior that underlie this mechanism, considering various scenarios such as binary choices or multi-choice problems. Through rigorous analysis, simulations, and real-world datasets, we hope to derive insights that explain why SC often leads to consistent improvement and under what conditions it might fail.

%The section that follows delves into the mathematical framework of majority voting, starting with modeling the problem as a binomial distribution and progressively generalizing to multinomial and Dirichlet distributions. 
The following section explores the mathematical explanation of SC, beginning with a simple binomial distribution model and gradually extending to more complex multinomial and Dirichlet distributions. By understanding the mathematical characteristics of these distributions, we hope to explain the empirical results observed in real-world reasoning tasks, thereby contributing to the ongoing efforts to harness the power of SC in a wide range of applications.

\begin{figure*}[t!]

  \centering
  \begin{subfigure}{.33\textwidth}
    \centering
    \includegraphics[width=\linewidth]{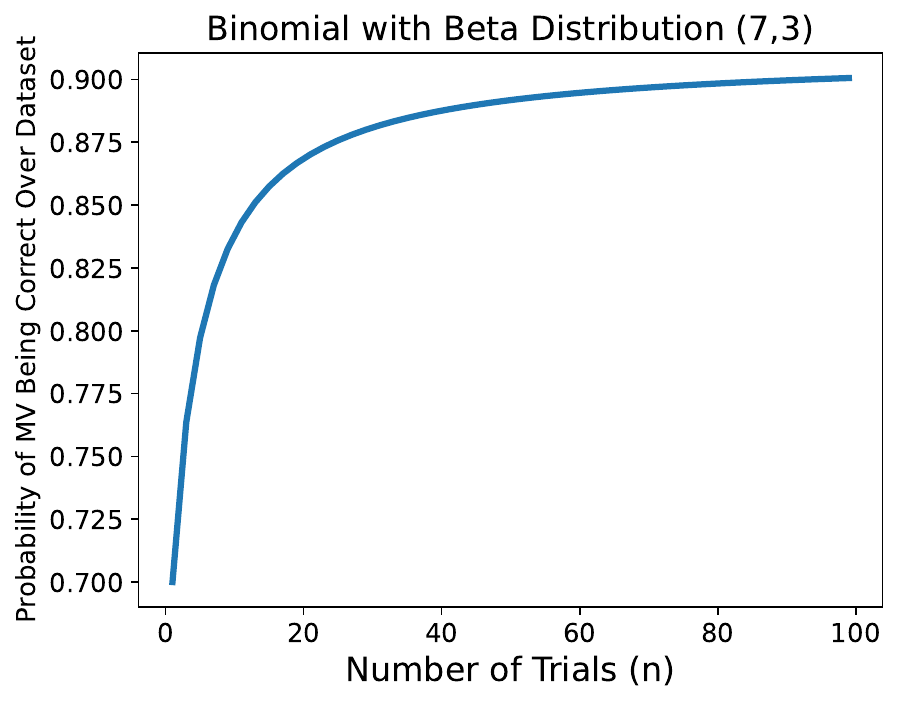}
    \caption{}
    \label{fig:binomial_beta_7_3}
  \end{subfigure}\hfill
  \begin{subfigure}{.33\textwidth}
    \centering
    \includegraphics[width=\linewidth]{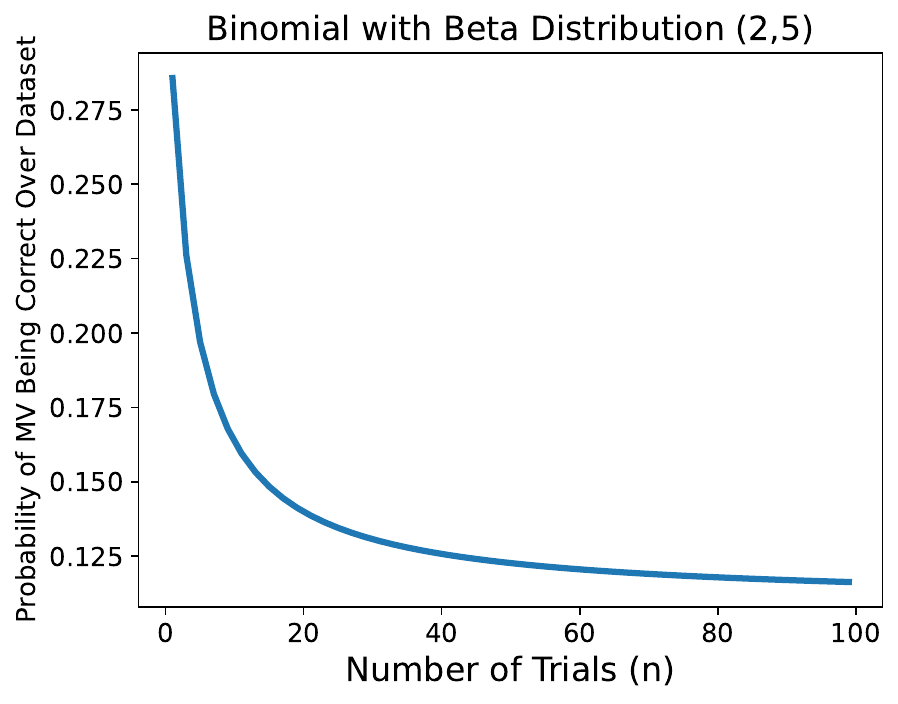}
    \caption{}
    \label{fig:binomial_beta_2_5}
  \end{subfigure}
    \begin{subfigure}{.32\textwidth}
    \centering
    \includegraphics[width=\linewidth]{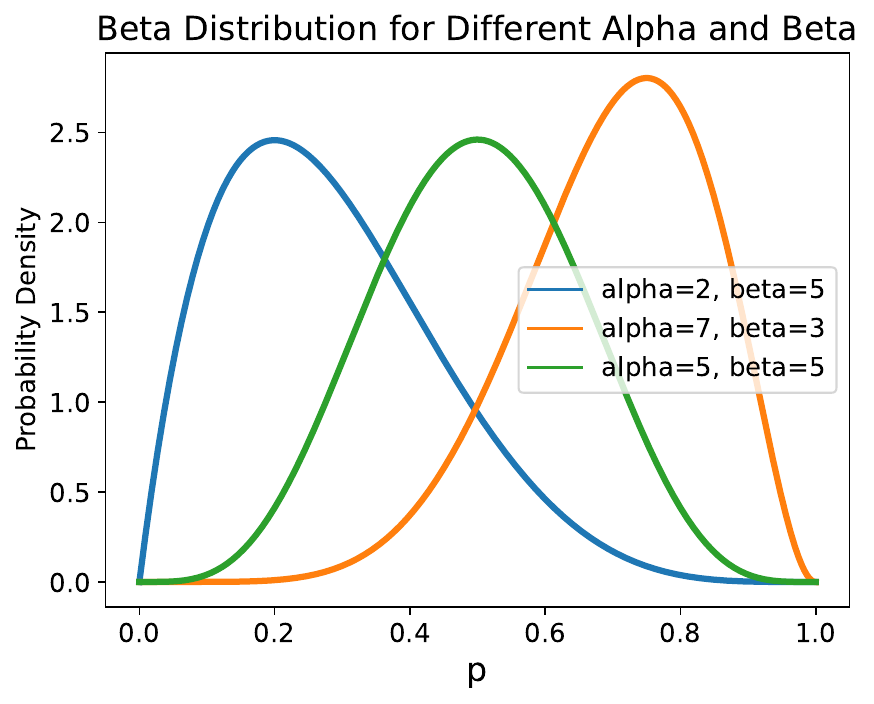}
    \caption{}
    \label{fig:beta_dist}
  \end{subfigure}
  \caption{Convergence of self consistency under different beta distributions. Here, a Beta distribution that peaks at high $p$ indicates that there are a lot of data examples where the model can solve with high probabilities, which leads to higher average self-consistency scores.
  } 
\label{fig:beta_distribution}
\end{figure*}

%%%%%%%%%%%%%%%%%%%%%%%%%%%%%%%%%%%%%%%%%
\subsection{Self-Consistency Results on Reasoning Tasks}

In our exploration of SC strategies applied to reasoning tasks, we conducted several experiments to analyze the effectiveness and behavior of different approaches. Figure~\ref{fig:budget_queries_vs_tokens} and Appendix \ref{appendix:more_models} illustrate our findings, including the results for different tasks.

% \begin{figure}[ht]
%   \centering
%   \fbox{
%     \begin{minipage}{.8\textwidth}
%       \centering
%       Placeholder for Majority Voting Plot on Multiple Tasks
%     \end{minipage}
%   }
%   \caption{Majority voting results on different reasoning tasks. Task A and Task B are represented in the same plot, showcasing the various effects of majority voting.}
%   \label{fig:majority_voting_results}
% \end{figure}

 The convergence patterns and the improvement as the number of trials increases are shown for each task, highlighting the impact of SC.

These visualizations demonstrate the potential of SC in enhancing reasoning tasks, leading to more robust and accurate solutions. In this section, we will provide a theoretical framework that could explain the gains from SC. Note that we use Self-Consistency (SC) and Majority-Vote (MV) interchangeably.

\subsection{Binomial}
We seek to analyze the behavior of parallel sampling with $n$ trials with self-consistency or SC. In this setup, given a set of problems $\{ x_i \}$, each problem's answer prediction (whether it is correct or not) can be modeled as a binomial distribution, assuming two choices (yes or no). Mathematically, the probability mass function for each problem's answer is given by:
\begin{equation}
f(X_i = k) = \binom{n}{k} p_i^k (1-p_i)^{n-k},
\end{equation}
where $X_i$ corresponds to the correct answer of the binomial distribution and $p_i$ represents the probability of a correct answer for the $i$-th problem.

We can calculate the probability that SC yields the correct solution over $n$ trials by calculating the probability that $X_i$ yields a value that is at least $n/2$. This is expressed as:
\begin{equation}
P(\text{{MV correct}} | x_i) = \sum_{k=\lceil n/2 \rceil}^n \binom{n}{k} p_i^k (1-p_i)^{n-k}.
\end{equation}

By plotting the probability of MV being correct as a function of $n$, we observe that as $n$ increases, $P(\text{{MV correct}} | x_i)$ either goes to 0 or 1, depending on whether $p_i > 0.5$ or $p_i < 0.5$ for this particular problem. This is evident in the synthetic experiment shown in Figure \ref{fig:prob_mv_correct_one_problem}.

If $p_i$ is extreme (closer to 1 or 0), then the convergence is fast, and the probability function can be described as:
\begin{equation}
\lim_{n \to \infty} P(\text{{MV correct}} | x_i) =
\begin{cases}
1 & \text{if } p_i > 0.5, \\
0 & \text{if } p_i < 0.5.
\end{cases}
\end{equation}

\begin{figure}[h]
  \centering
  \includegraphics[width=0.45\textwidth]{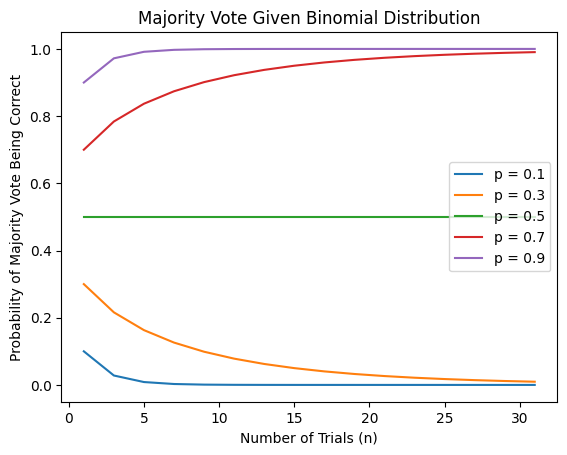}
  \caption{Probability of self consistency being correct for a given problem with varying $p$.}
  \label{fig:prob_mv_correct_one_problem}
\end{figure}

\begin{figure*}[ht]
  \centering
  \begin{subfigure}{.49\textwidth}
    \centering
    \includegraphics[width=\linewidth]{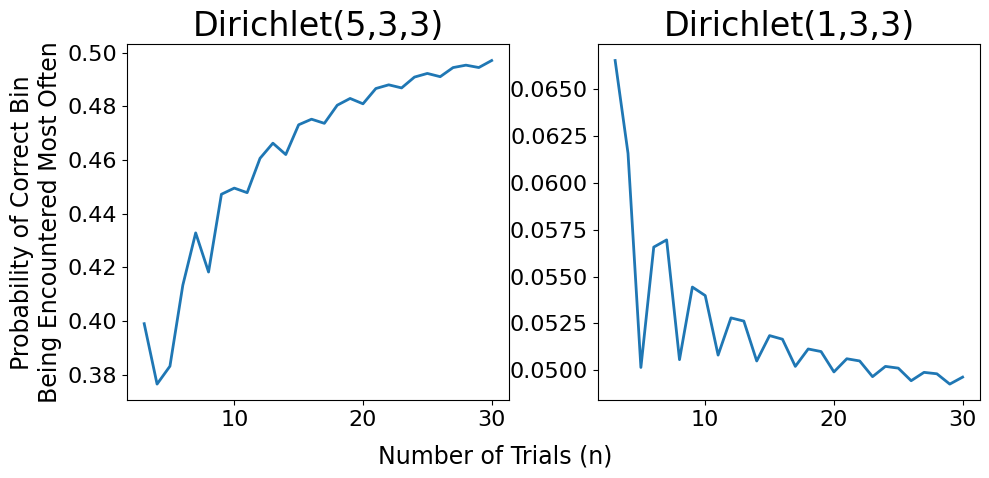}
    \caption{}
    \label{fig:plot1}
  \end{subfigure}
    \begin{subfigure}{.5\textwidth}
    \centering
    \includegraphics[width=\linewidth]{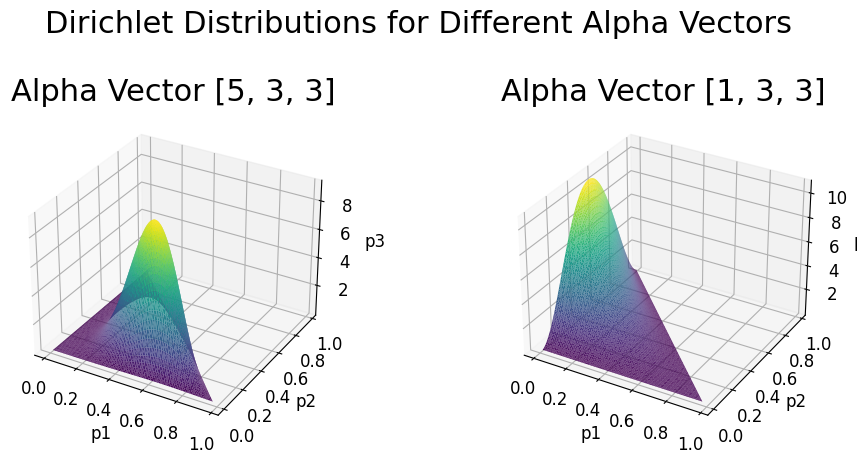}
    \caption{}
    \label{fig:plot3}
  \end{subfigure}
  \caption{Convergence of self consistency under different Dirichlet distributions with $K=3$} 
\label{fig:dirichlet_distribution}
\end{figure*}

On the other hand, if $p_i$ is close to 0.5, the convergence is slow, reflecting the uncertainty associated with an answer that is nearly equally likely to be correct or incorrect.

Over the set of all problems we consider, we place a beta distribution over $p_i$ and integrate $P(\text{{MV correct}} | x_i)$ over the set of all problems to obtain $P(\text{{MV correct}})$. This can be expressed mathematically as:
\begin{multline}
P(\text{{MV correct}}) \\ = \int_0^1 P(\text{{MV correct}} | p_i) \cdot f(p_i | \alpha, \beta) \, dp_i,
\end{multline}
where $f(p_i | \alpha, \beta)$ is the probability density function of the beta distribution with parameters $\alpha$ and $\beta$.

If we select a beta distribution where the mode peaks beyond 0.5, then we find that $P(\text{{MV correct}})$ increases as a function of $n$, albeit to a value less than 1 as you can see in Figure \ref{fig:beta_distribution}. This behavior explains our observation in real datasets directly.

This also implies that for datasets where majority vote leads to consistent improvement, the distribution of $p_i$ needs to be peaked greater than 0.5. There would also exist a set of problems where self-consistency leads to lowered performance, specifically for the set of problems where $p_i < 0.5$.

By carefully selecting the parameters of the beta distribution, we can control the characteristics of the majority voting process and gain insights into the behavior of parallel sampling across various datasets. This mathematical framework provides a powerful tool for understanding and optimizing the majority vote process in practical applications.

\subsection{Generalization to multinomial} 
We can further generalize this setup by considering each problem as being modeled by a multinomial distribution with $K$ choices. In this more generalized scenario, the distribution of probabilities over problems can also be modeled by a Dirichlet distribution.

Let $p = (p_1, p_2, \ldots, p_K)$ be the probabilities associated with the $K$ choices, and let $\alpha = (\alpha_1, \alpha_2, \ldots, \alpha_K)$ be the parameters of the corresponding Dirichlet distribution. The probability of obtaining a correct majority vote for a given problem is then:

\begin{equation}
P(\text{{MV correct}} | p) = \sum_{k=\lceil n/2 \rceil}^{n} \text{{multinomial}}(k; n, p),
\end{equation}
where the sum is taken over all combinations of $k$ votes that would result in a majority for the correct choice.

The overall probability of obtaining a correct majority vote, integrating over all problems, can be expressed as:

\begin{equation}
P(\text{{MV correct}}) = \int P(\text{{MV correct}} | p) \cdot f(p | \alpha) \, dp,
\end{equation}
where $f(p | \alpha)$ is the probability density function, which can be modeled by the Dirichlet distribution.

Following a similar simulation to the binary case, we find that the conclusions hold (see Figure \ref{fig:dirichlet_distribution}). Specifically, if the mode of the Dirichlet distribution is biased towards the correct choices, the probability of the majority vote being correct increases with $n$, and the set of problems where self-consistency leads to lowered performance can be characterized by the subset where the correct choice probabilities are below certain thresholds.

This generalization to multinomial and Dirichlet distributions adds complexity but also additional flexibility in modeling the majority voting process, making it applicable to a broader range of practical scenarios.

\clearpage

\newpage

% \clearpage
\begin{figure}[h]
\begin{center}
\includegraphics[width=0.4\textwidth]{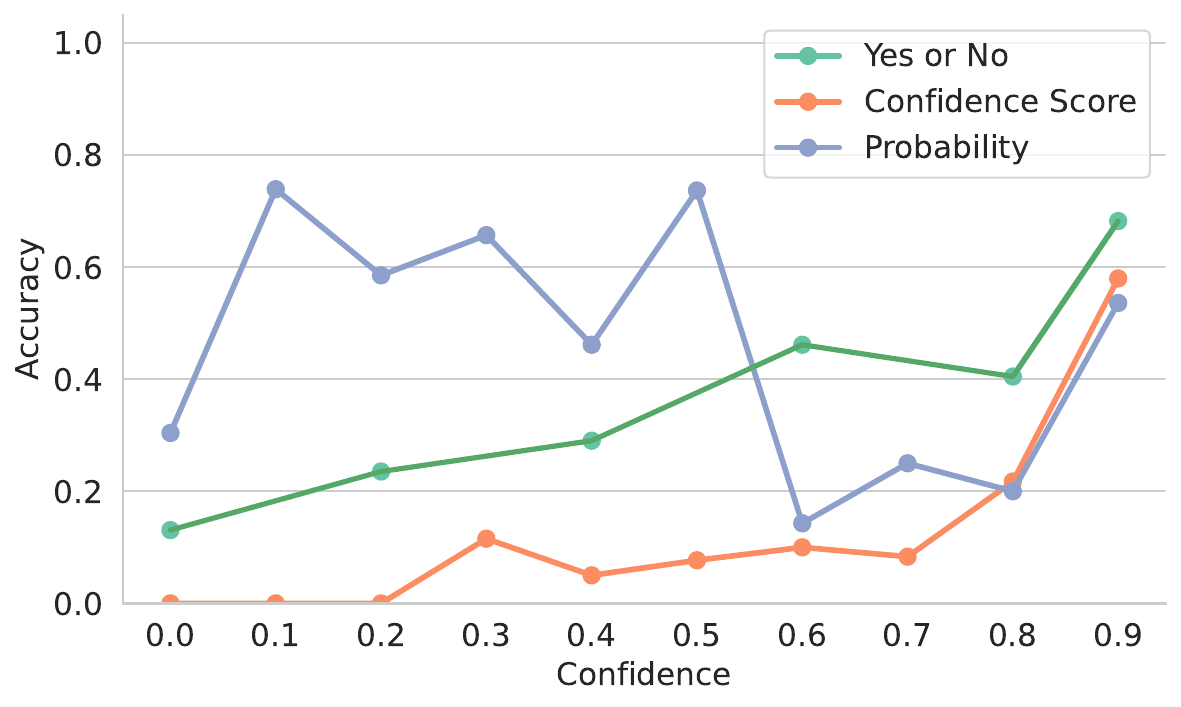}
\end{center}
\caption{Calibration result for the math reasoning datasets. Three different self-evaluation methods are calibrated here. 
} \label{fig:all_calibrations} 
\label{fig:self_eval_calibration_MATH}
\end{figure}

\begin{table}[h!]
\centering
\resizebox{\columnwidth}{!}{%
\begin{tabular}{c c c c } 
 \toprule
 Method & Correct Accuracy & Incorrect Accuracy & Total Accuracy \\ [0.5ex] 
 \midrule
 Yes or NO & 0.911 & 0.461 & 0.707 \\ 
 Score 1-10 & 0.995 & 0.149 & 0.613 \\
 Probability 0.0-1.0 &  0.886 & 0.115 & 0.537 \\
 \bottomrule
\end{tabular}
}
\caption{Self-evaluation accuracy on MATH with three methods}
\label{table:self_eval_accuracy_MATH}
\end{table}

\section{Self-Evaluation} \label{sec: self-evaluation}
\subsection{Self-Evaluation Method}
Given an answer, there are multiple ways we can prompt the LLM to evaluate that answer. Here we examine 3 possibilities for self-evaluation

\begin{enumerate}
    \item \textbf{Binary}\footnote{We also investigate a variant where we ask the model to think step by step before evaluating. While we see a small increase in performance for such a strategy, it also necessitates a big increase in the token budget. Further analysis is in the Supplement.} - we ask the model to output Yes/No as to whether the answer is correct. We do this multiple times and take the fraction of times the model answers Yes as the confidence of the model in the answer.
    \item \textbf{Numerical confidence} - we ask the model to output a score between 1 and 10 to indicate its confidence in the answer. We do this multiple times and take the average as the confidence of the model in the answer.
    \item \textbf{Confidence probability} - similar to the previous strategy except now we prompt the model to output a confidence between 0.0 and 1.0 and average it.
    % \item \ben{compare with implicit probability from generation?}
\end{enumerate}
The evaluation result is shown in Figure \ref{fig:self_eval_calibration_MATH} and Table \ref{table:self_eval_accuracy_MATH}. The binary Yes or No is the most well calibrated.
\begin{figure*}[]

\begin{center}
\includegraphics[width=1\textwidth]{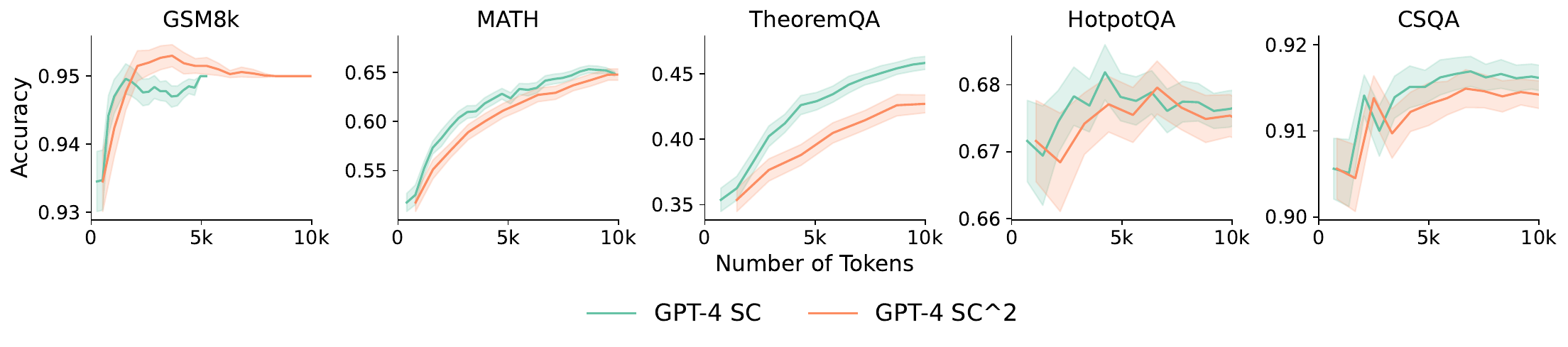}
\end{center}
\caption{$SC^2$ with total tokens being the budget. There are sizable improvements in using our method $SC^2$ on math reasoning tasks.
} \label{fig:SC_2_Num_Tokens} 
\label{fig:WCSC_All_datasets}
\end{figure*}
% \sid{@junlin Need plots here for GSM8K/MATH/TheoremQA}

\begin{figure*}[ht!]
  \centering
\includegraphics[width=1\textwidth]{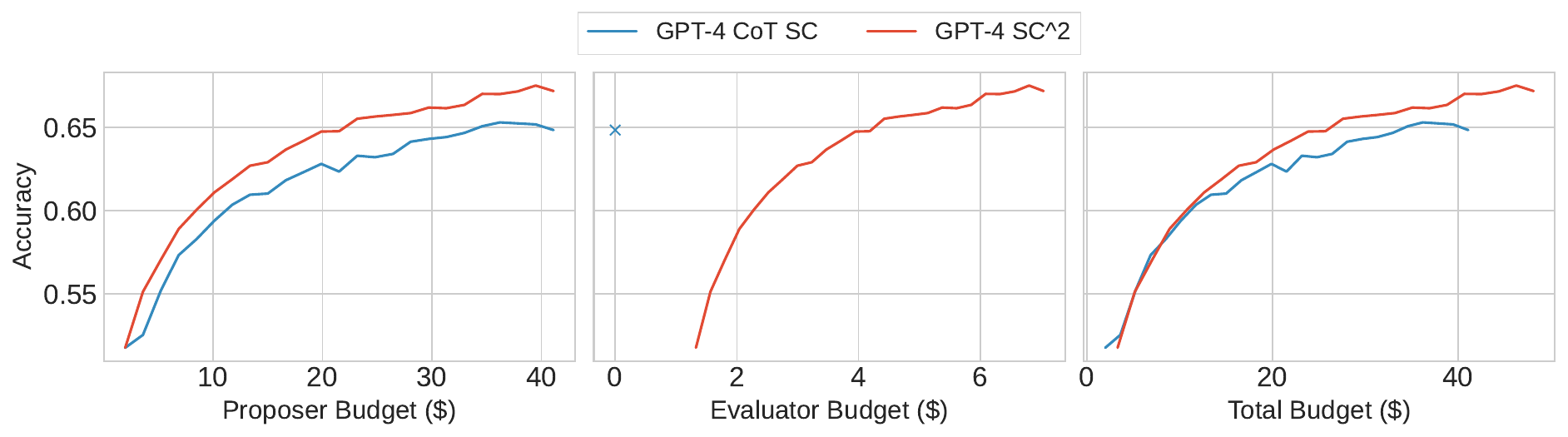}
  \caption{Separate proposer budget and evaluation budget on the dataset of MATH.}
  \label{fig:accuracy_vs_budget_WCSC_MATH}
\end{figure*}

\subsection{Self-evaluation is correlated with problem difficulty}
To get an understanding of whether models found it easier to evaluate answers to easier problems, we computed the following metric for a 100 problem subset of the GSM8K dataset. For each problem $i$, let $a_{ij}$ be the $j$th answer. We had 20 sampled answers per problem. We computed the fraction $c_i$ of answers that were correct. Our assumption was that $c_i$ indicates the difficulty of the problem -- the higher the value, the easier the problem. For each answer $a_{ij}$, we obtained the binary self-evaluation confidence as described in the beginning of this section (we sampled the evaluation 5 times). We then computed the correlation $\rho_i$ between the self-evaluation confidence for the answers $a_{ij}$ and the binary vector indicating whether the answers were correct or not. We then computed the correlation between $\rho_i$ and $c_i$. We obtained a correlation of 0.347 with a p-value of 0.00026 -- a clear indication that an increase in the problem difficulty results in the self-evaluation becoming more noisy. We repeated this experiment for MATH and TheoremQA and obtained correlations of 0.31 and 0.42 with p-values of 0.02 and 0.0025 respectively.

\subsection{Self-Confident Self-Consistency ($SC^2$) Details} \label{sec:self-evaluation:sc2}
 We take the answer which has the highest SC score as the predicted answer. Formally the definition is
\vspace*{-2mm}
\begin{align}
    SC^2_a = \sum_{a_i = a} \text{confidence}(a_i)
\end{align}
\vspace*{-2mm}

where $\text{confidence}(a_i) = \frac{\sum_{v_j} \mathbb{I}(v_j = \text{Yes})}{m}$ where $m$ denotes the number of Binary evaluations $v_j$ sampled. We apply this strategy to the MATH, TheoremQA (integer answer subset), TheoremQA (random subset), and HotpotQA datasets. $SC^2$ is consistently on par or better than a simple majority vote. The results are in Figure \ref{fig:WCSC_All_datasets}. $SC^2$ achieves non-trivial gain for math reasoning tasks but the overall costs increase quite a bit. This prompts us to inquire whether the achieved performance boost justifies the additional costs incurred. However, if we have the option to cache, then during self-evaluation, previous questions and answers can be cached and don't need to be encoded again. This can save a lot of budget and the new results would look like Figure \ref{fig:WCSC_All_datasets_cached}. We see non-trivial gains for the math reasoning datasets. However, for TheoremQA we see markedly smaller gains. We hypothesize that the reason for this is that TheoremQA is a harder dataset for the model. As we showed in the previous section, self-evaluation ability decreases as problem difficulty increases. GPT-4 shows a self-evaluation ability of no better than random for TheoremQA and thus we observe very small improvement. Overall, a budget-aware comparison of reasoning methods is a healthy approach to compare among vastly different methods.

\begin{table*}[t!]
\centering
\resizebox{\columnwidth}{!}{%
\begin{tabular}{||c c c c||} 
 \hline
 Method & Top1 & Best out of all & Total Accuracy \\ [0.5ex] 
 \hline
 ToT b=5 (GPT-4,GPT-4) & 0.74 & 0.76 & 0.4 \\ 
 ToT b=3 (GPT-4,GPT-4) & 0.77 & 0.77 & 0.49 \\
 ToT b=1 (GPT-4,GPT-4) & 0.65 & 0.65 & 0.65 \\
 ToT eval once (GPT-4,GPT-4) & 0.73 & 0.75 & 0.352 \\
 CoT 100 times (GPT-4) & 0.17 & 0.56 & 0.0756 \\ 
ToT Random Eval (GPT-4) & 0.0 & 0.04 & 0.008 \\ 
 ToT b=5 (GPT-3.5,GPT-3.5) & 0.25 & 0.35 & 0.11 \\ 
CoT 100 times (GPT-3.5) & 0.04 & 0.46 & 0.0252 \\ 
 ToT b=5 (GPT-4,GPT-3.5) & 0.68 & 0.72 & 0.302 \\ 
ToT b=5 (GPT-3.5,GPT-4) & 0.3 & 0.38 & 0.156 \\ 
 [1ex] 
 \hline
\end{tabular}
}
\caption{Various results on Game of 24. ToT refers to Tree-of-Thoughts. For ToT, the first model name in the parenthesis refers to the model used to \textit{generate} the candidate thoughts, while the second model name refers to the model used to \textit{evaluate} the candidate thoughts.}
\label{table:ToT_performances}
\end{table*}

\subsubsection{Budget-efficiency}

The strategy requires only a handful of extra tokens ($m$ additional tokens per answer corresponding to the Yes/No) to execute (Figure \ref{fig:accuracy_vs_budget_WCSC_MATH}). However, it does require more encoded tokens (We can sample all of the $m$ additional tokens as part of a single query). Thus if one is self-hosting the model, this strategy has only marginal additional cost.

\subsubsection{Query vs Token budget}

While we have discussed both query and token budget in this paper, token budget has some notable advantages as a metric. 

\paragraph{Theoretical aspects}
Equivalence in the number of queries can be arbitrarily far from the equivalence in the amount of compute.~\citet{merrill2023expresssive} and~\citet{perez2021attention} both show that the expressive power of transformers can be greatly enhanced by generating intermediate steps in the computation (colloquially called chain of thought).~\citet{merrill2023expresssive} shows that without any bound in the number of steps, an encoder-decoder architecture with only one encoder and three decoder layers can simulate a Turing Machine and thus a single query to such a Transformer can perform computations with arbitrarily large amount of compute.~\citet{perez2021attention} shows that even for decoder-only transformers, allowing for polynomial-sized chains of thought makes it powerful enough to do, in a single query, any computation a Turing Machine can do in polynomial time. While the number of queries metric fails to capture this, by contrast, the number of tokens metric which is novel to our paper, does capture this aspect as it by definition includes the length of the generated thought as part of the compute.
\begin{figure}[t]

  \centering
  \begin{subfigure}{0.48\textwidth}
    \centering
    \includegraphics[width=\linewidth]{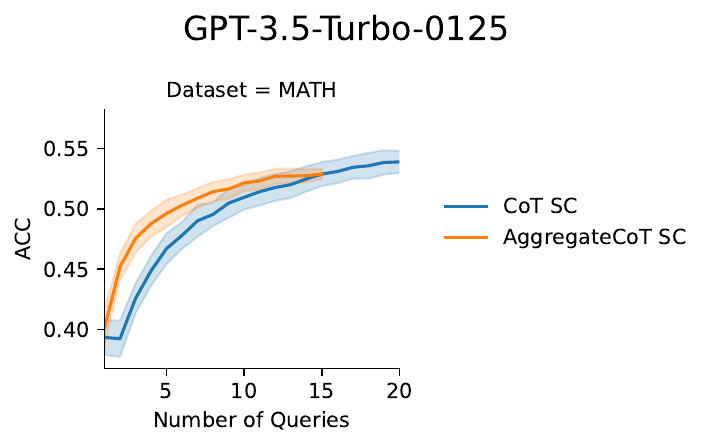}
    \caption{Query Budget}
    \label{fig:custome_reasoning1}
  \end{subfigure}
  \begin{subfigure}{.3\textwidth}
    % \centering
    \hspace*{-16mm}
    \includegraphics[width=\linewidth]{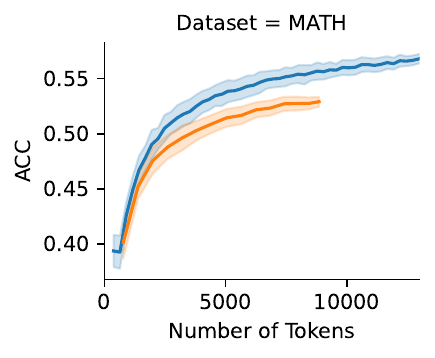}
    \caption{Token Budget}
    \label{fig:custome_reasoning2}
  \end{subfigure}
  \caption{We evaluated this custom reasoning strategy on MATH with GPT-3.5-Turbo-0125 for 15 queries, so in theory it should generate 15*4=60 responses. Here is the result based on the number of queries metric (we name the custom reasoning strategy AggregateCoT). We can find that it never outperforms Chain-of-Thought Self-Consistency with same amount of tokens. The "improvement" previously was an unfair comparision because the custome reasoning strategy will use much more tokens per query.
  } 
\label{fig:custome_reasoning}
\end{figure}
\paragraph{Practical aspects} The above is not just a theoretical consideration. In Figure \ref{fig:custome_reasoning1} we demonstrate, a custom reasoning strategy that at first glance, seems to outperform self-consistency – based on the number of queries metric. However, when we properly take the holistic compute budget into account via the number of tokens metric, we can see that self-consistency is more token-efficient (Figure \ref{fig:custome_reasoning2}). That is, the number of tokens as a metric of budget captures the nuances of resources required for LLM reasoning more properly.
\section{More Ablation Results for Tree of Thought}
In Table \ref{table:ToT_performances} you can see more ToT results on the task of Game of 24. Most of the results are shown in the Figure \ref{fig:accuracy_vs_budget_ToT_Game24}. The table mainly shows the ablation for when using GPT-3.5 as the proposer and GPT-4 as the evaluator. We see that the performance is better than using a GPT-3.5 as the evaluator but far below the performance of using GPT-4 as the proposer.
\section{Terms and Licenses}
GSM8K, MATH, TheoremQA, CSQA are under the MIT license. HotpotQA is under the CC BY-SA 4.0 License. All the datasets and models are used for their intended use.

\clearpage
\onecolumn
\section{Results From More Models}\label{appendix:more_models}
\subsection{MAD \& Reflexion}
Here we extend the results to a variety of models: GPT-3.5-Tubo-0125, Mistral-7B-Instruct-v0.2, LLaMA-2-70b-chat, and Mixtral-8x7B-Instruct-v0.1.
Overall, we find similar trends that self-consistency is extremely competitive compared to multi-agent debate and reflexion, when evaluated in a budget-aware manner.

We observed that it is very consistent that CoT with self-consistency beat other reasoning strategies across models with various sizes/training procedures. Multi-agent debate and Reflexion often decrease performances with more budget. This is not surprising considering our analysis in Section \ref{sec:Analysis}. Note that for LLaMA-2-70b-chat, we can't run Mad and Reflexion to the same amount of budget as CoT with self-consistency due to the context limit of around 4k. But the trend stays similar.

\begin{figure*}[h]
\begin{center}
% \vspace{-0.1cm}
\begin{subfigure}[b]{\textwidth}
\centering
\includegraphics[width=1\textwidth]{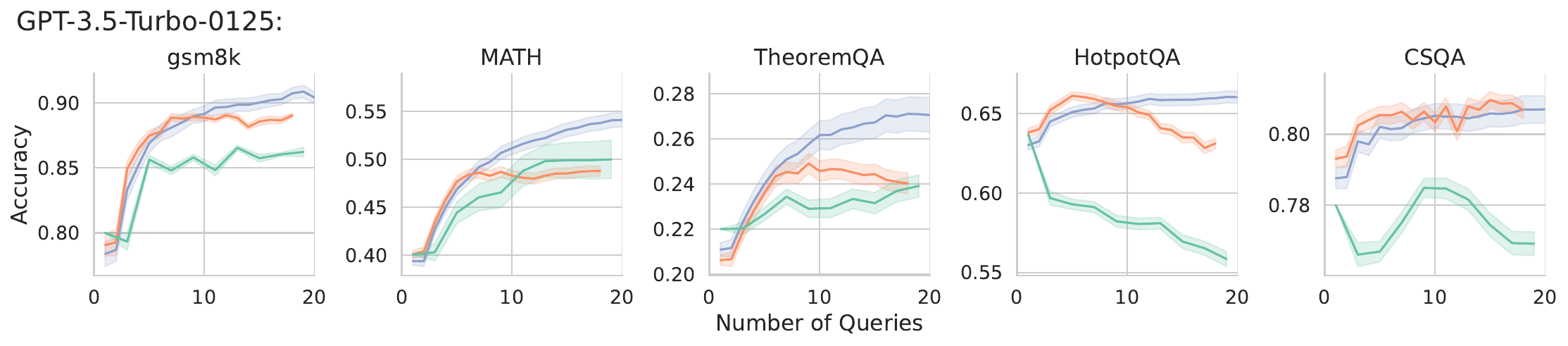}
\caption{} 
\end{subfigure}
\begin{subfigure}[b]{\textwidth}
\centering
\includegraphics[width=1\textwidth]{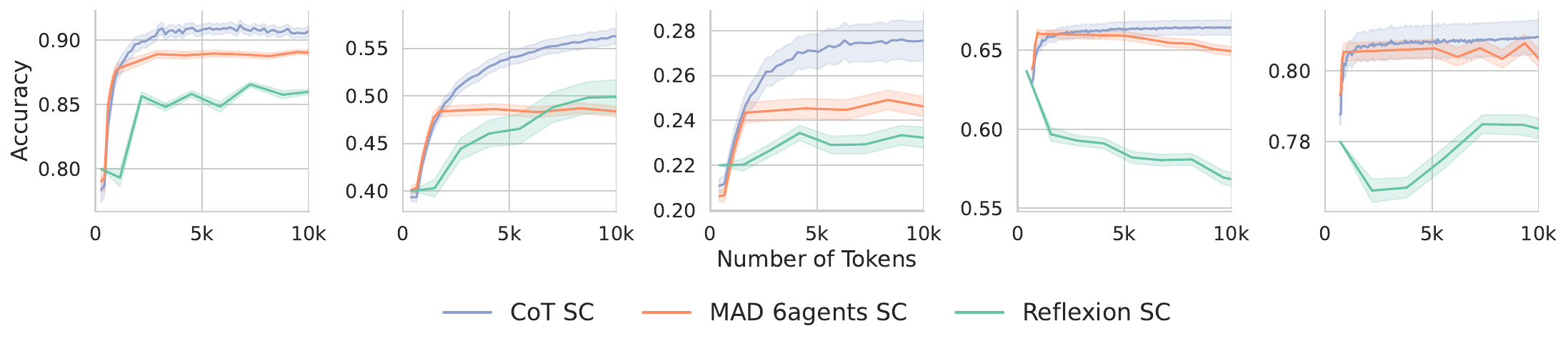}
\caption{ } 
\end{subfigure}
\end{center}
\vspace*{-3mm}
\caption{GPT-3.5-0125: (a) Performance@Number of Queries Plots for all 5 datasets. (b) Performance@Number of Tokens for all 5 datasets. } \label{fig:0125_budget_queries_vs_tokens}
\end{figure*}

\begin{figure*}[h]
\begin{center}
% \vspace{-0.1cm}
\begin{subfigure}[b]{\textwidth}
\centering
\includegraphics[width=1\textwidth]{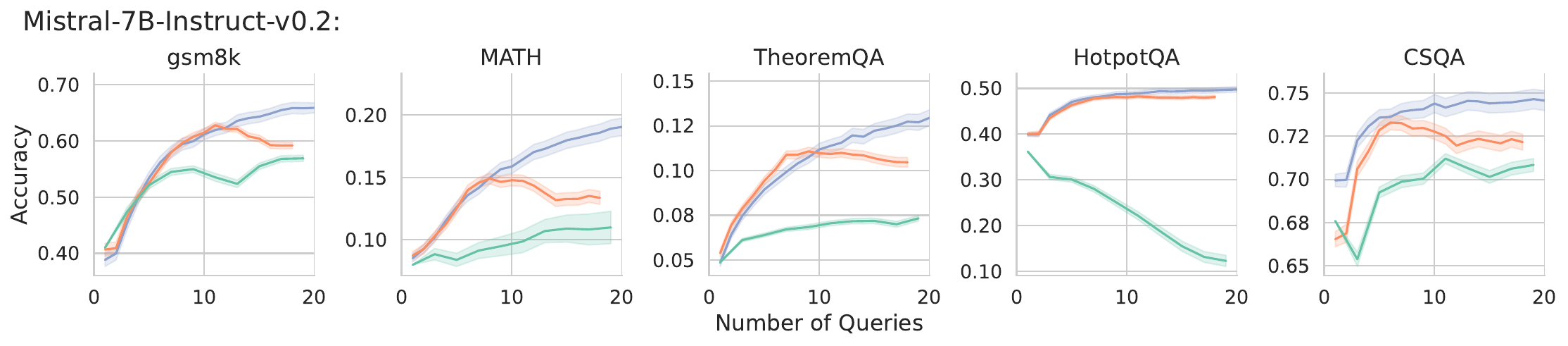}
\caption{} 
\end{subfigure}
\begin{subfigure}[b]{\textwidth}
\centering
\includegraphics[width=1\textwidth]{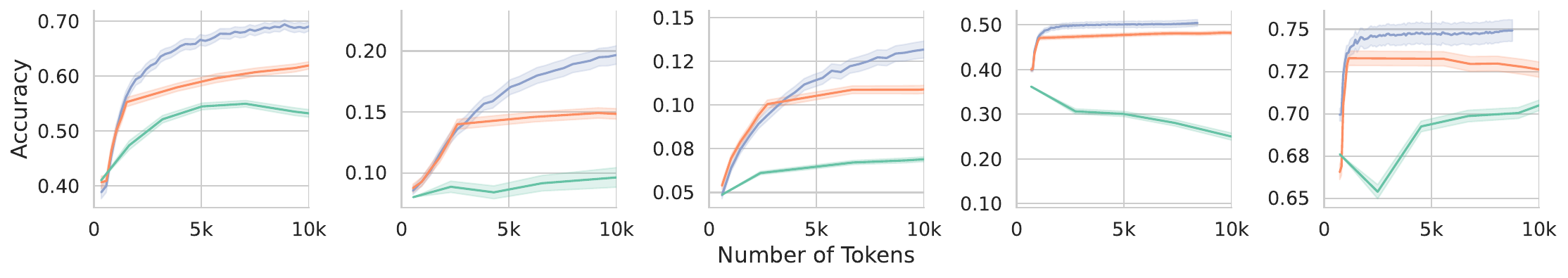}
\caption{ } 
\end{subfigure}
\end{center}
\vspace*{-3mm}
\caption{Mistral-7B-Instruct-v0.2: (a) Performance@Number of Queries Plots for all 5 datasets. (b) Performance@Number of Tokens for all 5 datasets. } \label{fig:mistral_MAD_budget_queries_vs_tokens}
\end{figure*}

\begin{figure*}[h]
\begin{center}
% \vspace{-0.1cm}
\begin{subfigure}[b]{\textwidth}
\centering
\includegraphics[width=1\textwidth]{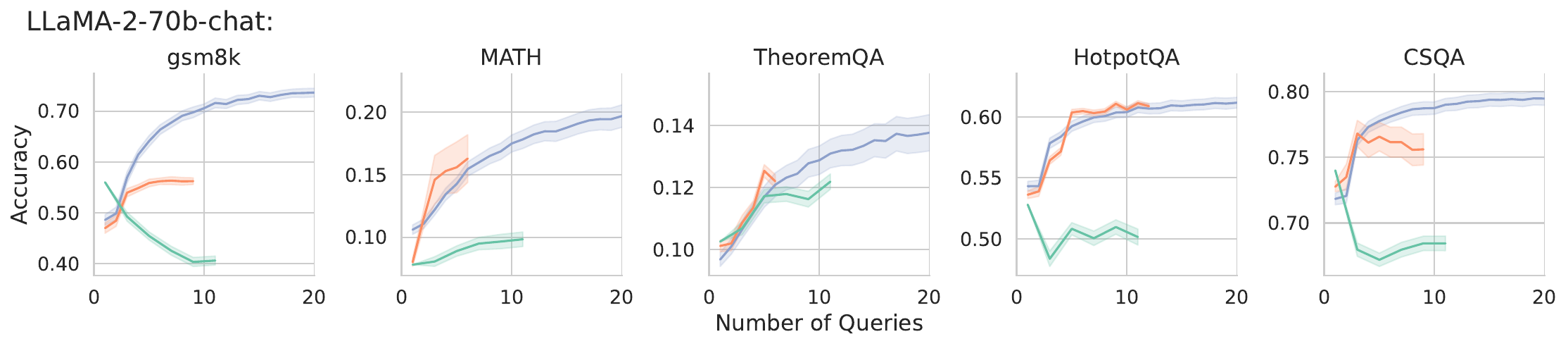}
\caption{} 
\end{subfigure}
\begin{subfigure}[b]{\textwidth}
\centering
\includegraphics[width=1\textwidth]{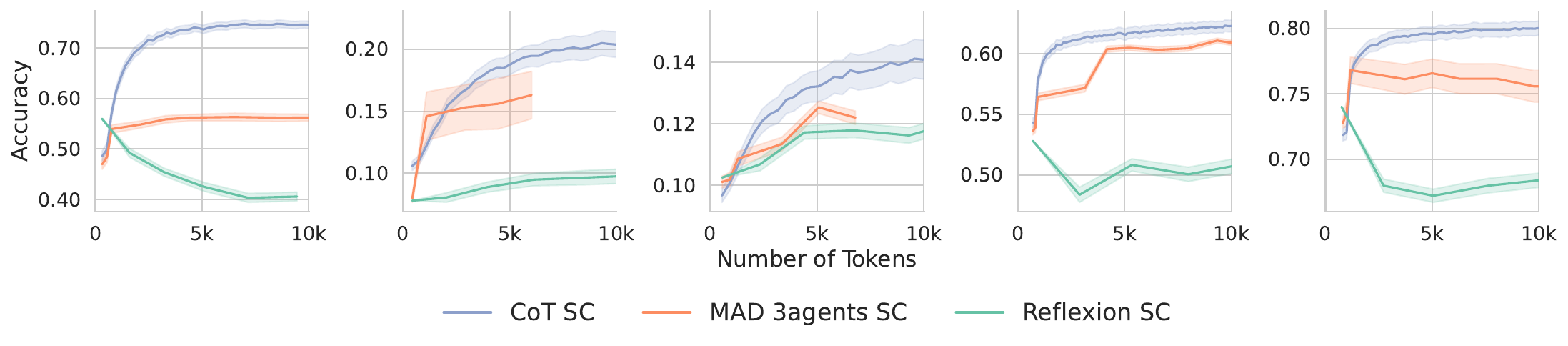}
\caption{ } 
\end{subfigure}
\end{center}
\vspace*{-3mm}
\caption{LLaMA-2-70b-chat: (a) Performance@Number of Queries Plots for all 5 datasets. (b) Performance@Number of Tokens for all 5 datasets. } \label{fig:llama_budget_queries_vs_tokens}
\end{figure*}

\begin{figure*}[h]
\begin{center}
% \vspace{-0.1cm}
\begin{subfigure}[b]{\textwidth}
\centering
\includegraphics[width=1\textwidth]{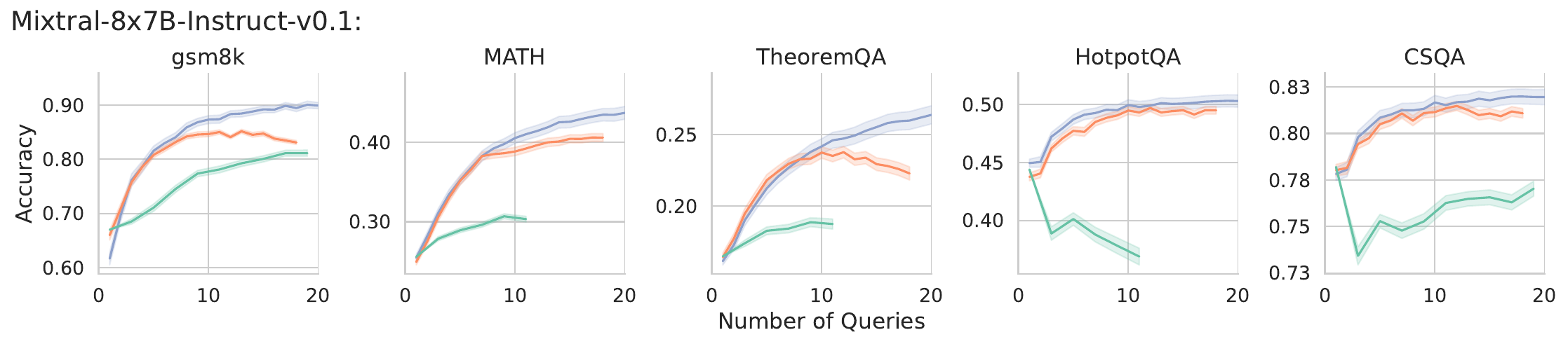}
\caption{}
\end{subfigure}
\begin{subfigure}[b]{\textwidth}
\centering
\includegraphics[width=1\textwidth]{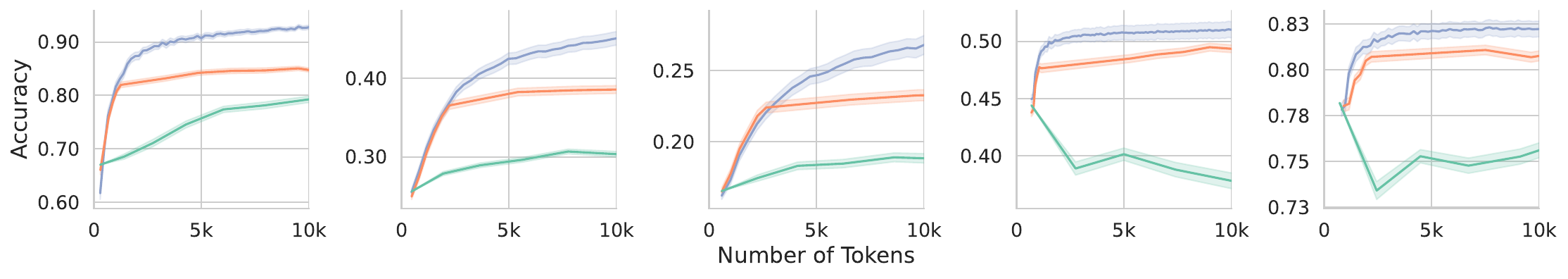}
\caption{ } 
\end{subfigure}
\end{center}
\vspace*{-3mm}
\caption{Mixtral-8x7B-Instruct-v0.1: (a) Performance@Number of Queries Plots for all 5 datasets. (b) Performance@Number of Tokens for all 5 datasets. } \label{fig:mixtral_budget_queries_vs_tokens}
\end{figure*}
\clearpage
\subsection{Three Other Reasoning Strategies} \label{appendix:more_models:three_other_strategies}

In this section, we will evaluate on three other reasoning strategies in the self-consistency family: Plan and Solve \cite{wang2023planandsolve}, Least to Most Prompting \cite{zhou2022leasttomost}, and Progressive Hint Prompting \cite{zheng2023progressivehint}. 

\paragraph{Plan and Solve} It asks LLMs to do some planning before solving a question. It is like an extension to CoT.

\paragraph{Least to Most Prompting} This strategy prompts the model to decompose a question first and then answer each subquestion before aggregating them to the final answer.

\paragraph{Progressive Hint Prompting} This strategy uses previous answers as hints to generate next answer.

All three new strategies here can be integrated with self-consistency seamlessly, since they are mostly just variants of chain-of-thought. Based on the plots, it seems that normal self-consistency is still very competitive, but different prompting styles can make a big difference. For some models and some datasets, a strategy other than CoT converges to a higher performance. This is strong evidence that self-consistency is a really budget-effective strategy.

\begin{figure*}[h]
\begin{center}
% \vspace{-0.1cm}
\begin{subfigure}[b]{\textwidth}
\centering
\includegraphics[width=1\textwidth]{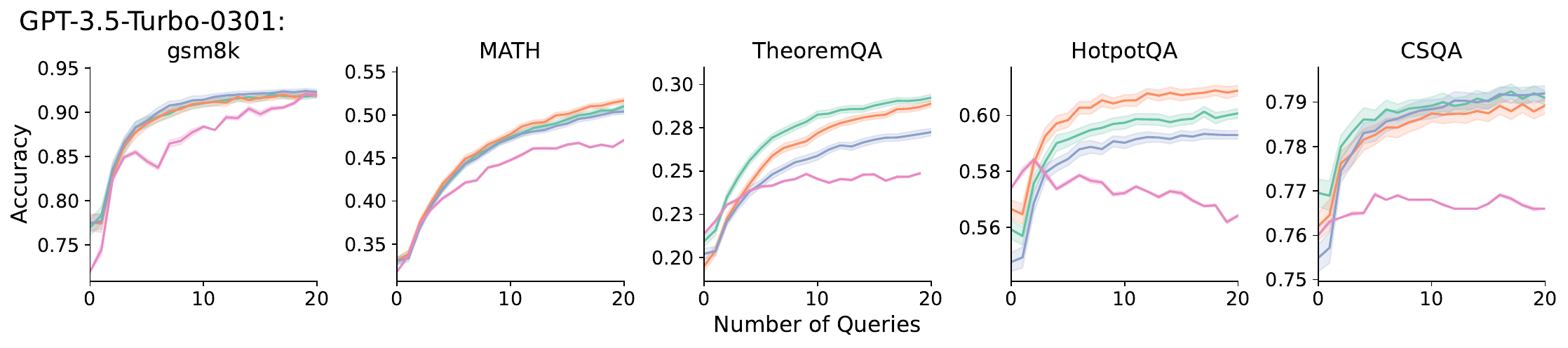}
\caption{} 
\end{subfigure}
\begin{subfigure}[b]{\textwidth}
\centering
\includegraphics[width=1\textwidth]{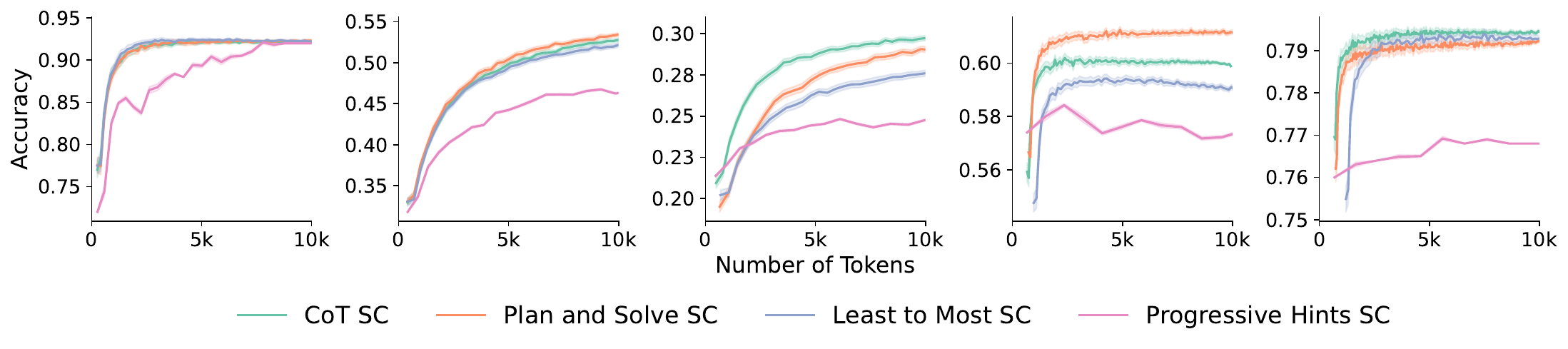}
\caption{ } 
\end{subfigure}
\end{center}
\vspace*{-3mm}
\caption{GPT-3.5-0301: (a) Performance@Number of Queries Plots for all 5 datasets. (b) Performance@Number of Tokens for all 5 datasets. } \label{fig:0301_new_budget_queries_vs_tokens}
\end{figure*}

\begin{figure*}[h]
\begin{center}
% \vspace{-0.1cm}
\begin{subfigure}[b]{\textwidth}
\centering
\includegraphics[width=1\textwidth]{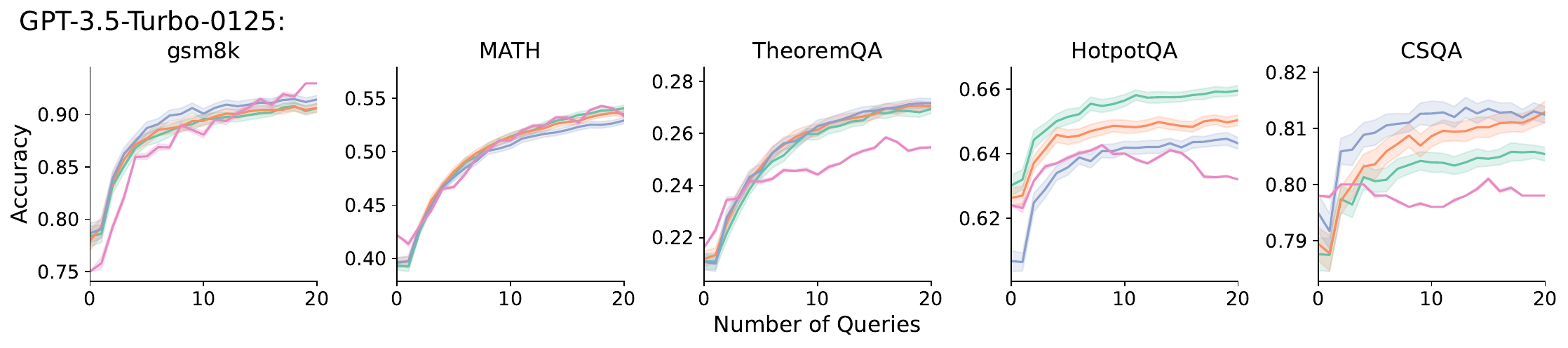}
\caption{} 
\end{subfigure}
\begin{subfigure}[b]{\textwidth}
\centering
\includegraphics[width=1\textwidth]{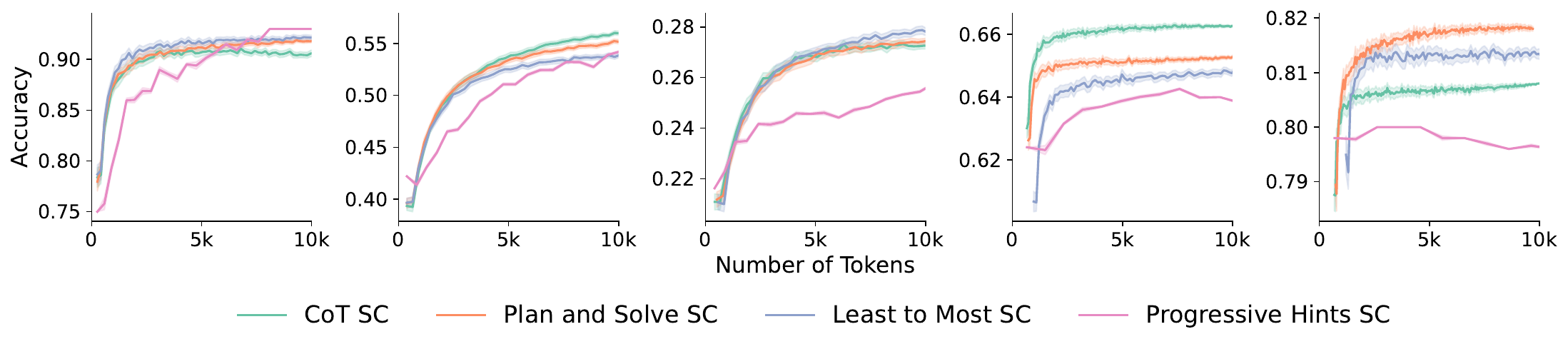}
\caption{ } 
\end{subfigure}
\end{center}
\vspace*{-3mm}
\caption{GPT-3.5-0125: (a) Performance@Number of Queries Plots for all 5 datasets. (b) Performance@Number of Tokens for all 5 datasets. } \label{fig:0125_new_budget_queries_vs_tokens}
\end{figure*}

\begin{figure*}[h]
\begin{center}
% \vspace{-0.1cm}
\begin{subfigure}[b]{\textwidth}
\centering
\includegraphics[width=1\textwidth]{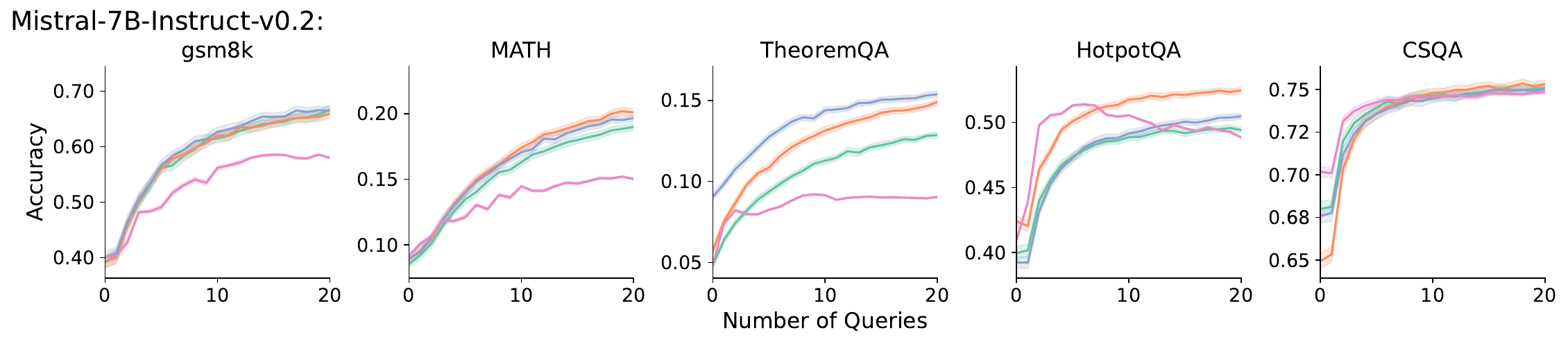}
\caption{} 
\end{subfigure}
\begin{subfigure}[b]{\textwidth}
\centering
\includegraphics[width=1\textwidth]{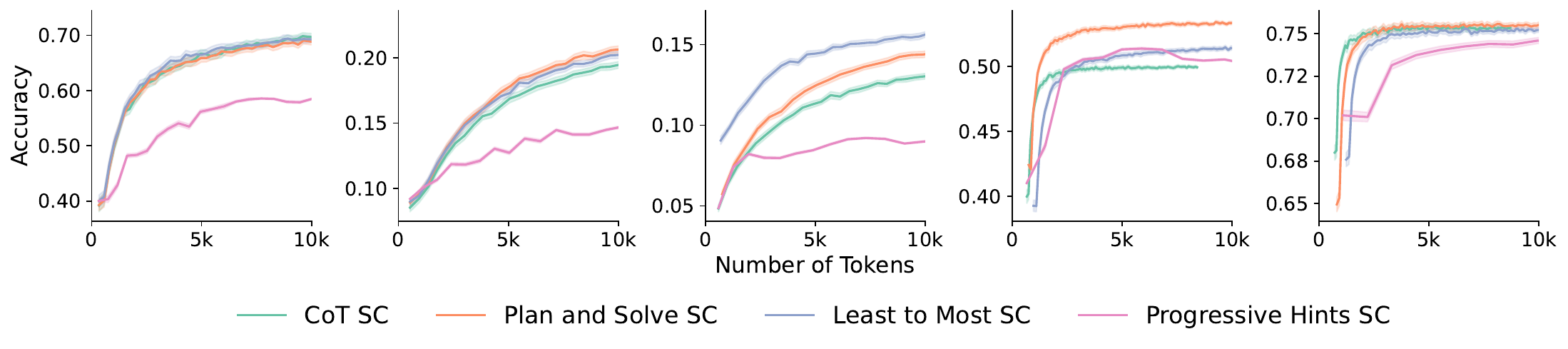}
\caption{ } 
\end{subfigure}
\end{center}
\vspace*{-3mm}
\caption{Mistral-7B-Instruct-v0.2: (a) Performance@Number of Queries Plots for all 5 datasets. (b) Performance@Number of Tokens for all 5 datasets. } \label{fig:mistral_budget_queries_vs_tokens}
\end{figure*}

\begin{figure*}[h]
\begin{center}
% \vspace{-0.1cm}
\begin{subfigure}[b]{\textwidth}
\centering
\includegraphics[width=1\textwidth]{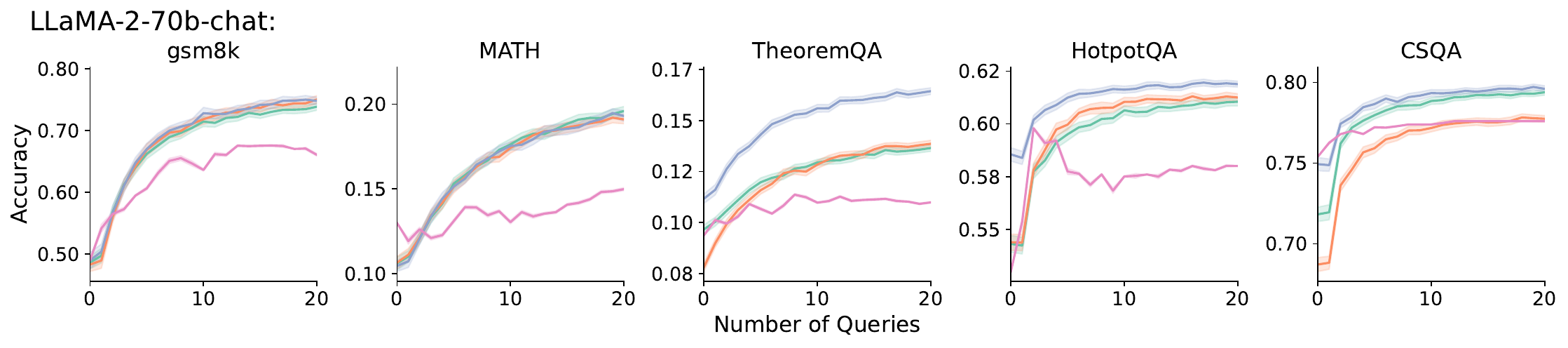}
\caption{} 
\end{subfigure}
\begin{subfigure}[b]{\textwidth}
\centering
\includegraphics[width=1\textwidth]{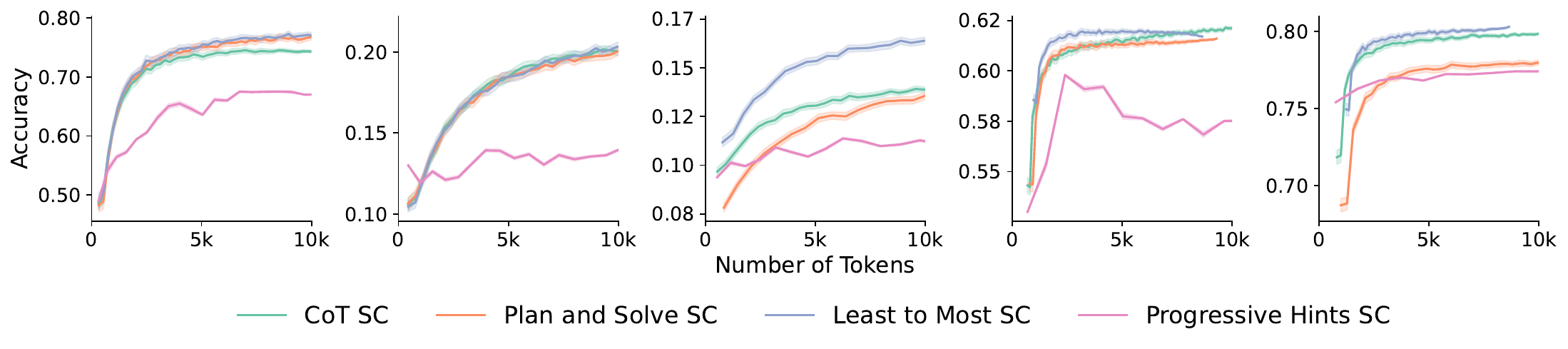}
\caption{ } 
\end{subfigure}
\end{center}
\vspace*{-3mm}
\caption{LLaMA-2-70b-chat: (a) Performance@Number of Queries Plots for all 5 datasets. (b) Performance@Number of Tokens for all 5 datasets. } \label{fig:llama_new_budget_queries_vs_tokens}
\end{figure*}

\begin{figure*}[h]
\begin{center}
% \vspace{-0.1cm}
\begin{subfigure}[b]{\textwidth}
\centering
\includegraphics[width=1\textwidth]{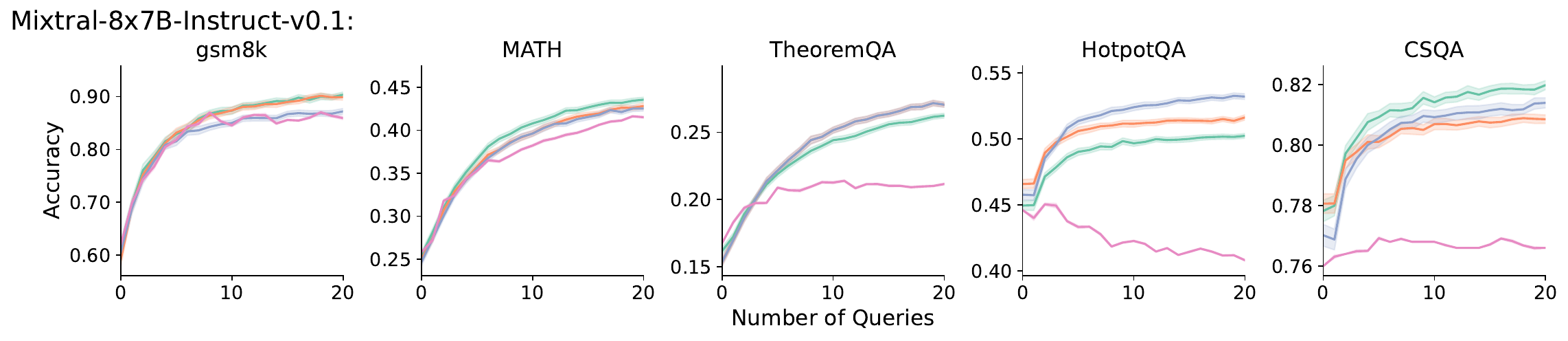}
\caption{} 
\end{subfigure}
\begin{subfigure}[b]{\textwidth}
\centering
\includegraphics[width=1\textwidth]{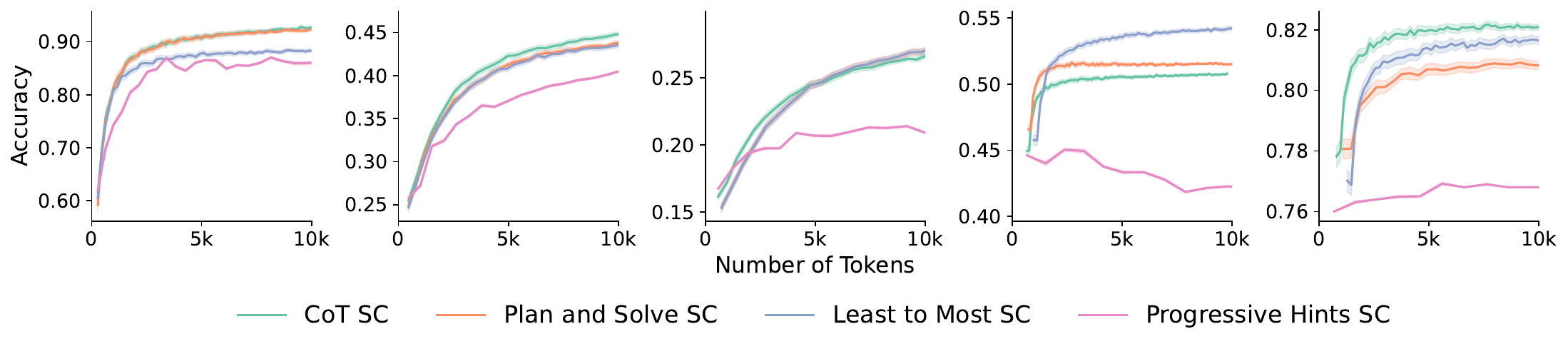}
\caption{ } 
\end{subfigure}
\end{center}
\vspace*{-3mm}
\caption{Mixtral-8x7B-Instruct-v0.1: (a) Performance@Number of Queries Plots for all 5 datasets. (b) Performance@Number of Tokens for all 5 datasets. 
} \label{fig:mixtral_new_budget_queries_vs_tokens}
\end{figure*}

\end{document}